\documentclass[10pt,journal,compsoc]{IEEEtran}
\ifCLASSOPTIONcompsoc
  \usepackage[nocompress]{cite}
\else
  \usepackage{cite}
\fi
\ifCLASSINFOpdf
\else
\fi
\usepackage{booktabs}
\usepackage{threeparttable}
\usepackage{multirow}
\usepackage{setspace}
\usepackage{subfig,caption}
\usepackage{graphicx}
\usepackage[colorlinks,linkcolor=black,anchorcolor=black,citecolor=black,urlcolor=black]{hyperref}

\usepackage{amsmath}
\usepackage{amssymb}

\usepackage{algorithmic}

\usepackage{array}
\usepackage{url}

\usepackage[switch]{lineno} 

\begin{document}
%
\title{Graphical Modeling for Multi-Source \\ Domain Adaptation}
%
%
%
%

\author{Minghao~Xu\textsuperscript{\rm *},
        Hang~Wang\textsuperscript{\rm *},
        Bingbing~Ni\IEEEauthorrefmark{2} \\
        \textsuperscript{\rm *}{\small equal contribution} \quad
	\textsuperscript{\rm $\dagger$}{\small corresponding author} \\
\IEEEcompsocitemizethanks{\IEEEcompsocthanksitem All authors are with Shanghai Jiao Tong University, Shanghai 200240, China. H. Wang and B. Ni are also with Huawei Hisilicon. \protect \\
E-mail: \{xuminghao118, Wang-\,-Hang, nibingbing\}@sjtu.edu.cn
\IEEEcompsocthanksitem \textsuperscript{\rm *} Authors contributed equally to this work.
\IEEEcompsocthanksitem \IEEEauthorrefmark{2} Corresponding author: Bingbing Ni.
}
}

\IEEEtitleabstractindextext{%
\begin{abstract}

Multi-Source Domain Adaptation (MSDA) focuses on transferring the knowledge from multiple source domains to the target domain, which is a more practical and challenging problem compared to the conventional single-source domain adaptation. In this problem, it is essential to model multiple source domains and target domain jointly, and an effective domain combination scheme is also highly required. The graphical structure among different domains is useful to tackle these challenges, in which the interdependency among various instances/categories can be effectively modeled. In this work, we propose two types of graphical models, \emph{i.e.} \textbf{C}onditional \textbf{R}andom \textbf{F}ield for MSDA (\emph{CRF-MSDA}) and \textbf{M}arkov \textbf{R}andom \textbf{F}ield for MSDA (\emph{MRF-MSDA}), for cross-domain joint modeling and learnable domain combination. In a nutshell, given an observation set composed of a query sample and the semantic prototypes (\emph{i.e.} representative category embeddings) on various domains, the CRF-MSDA model seeks to learn the joint distribution of labels conditioned on the observations. We attain this goal by constructing a relational graph over all observations and conducting local message passing on it.
By comparison, MRF-MSDA aims to model the joint distribution of observations over different Markov networks via an energy-based formulation, and it can naturally perform label prediction by summing the joint likelihoods over several specific networks. Compared to the CRF-MSDA counterpart, the MRF-MSDA model is more expressive and possesses lower computational cost. We evaluate these two models on four standard benchmark data sets of MSDA with distinct domain shift and data complexity, and both models achieve superior performance over existing methods on all benchmarks. In addition, the analytical studies illustrate the effect of different model components and provide insights about how the cross-domain joint modeling performs. Our code is available at \url{https://github.com/Francis0625/Graphical-Modeling-for-Multi-Source-Domain-Adaptation}.


\end{abstract}

}

\maketitle
\IEEEdisplaynontitleabstractindextext

\ifCLASSOPTIONpeerreview
\begin{center} \bfseries EDICS Category: 3-BBND \end{center}
\fi
%
\IEEEpeerreviewmaketitle

\IEEEraisesectionheading{\section{Introduction}\label{sec:introduction}}

%
%
%
%
\IEEEPARstart{T}{he} Unsupervised Domain Adaptation (UDA) methods~\cite{dan,revgrad,deepcoral,adda,cdan,max_discrepancy,larger_norm} assume a single source domain with supervision and aim to transfer the knowledge acquired from the source domain to another unsupervised target domain. However, in real-world applications, it is unreasonable to assume that the labeled data are drawn from a single data distribution. Actually, these samples are always collected from different deployment environments, \emph{i.e.} from multiple domains. For example, in an object classification task, one may have access to the annotated images captured in the morning, afternoon and evening, respectively, and the objective is to categorize the images captured at dawn. In addition, the diversity of weather, illumination and backgrounds can all lead to the existence of multiple domains in a specific data set. The problem under such a scenario is known as \emph{Multi-Source Domain Adaptation} (MSDA)~\cite{msda_theory_nips08}, in which one seeks to boost the model's performance on target domain by integrating the transferrable knowledge from various source domains. By employing MSDA algorithms' power of aligning multiple domains, we can better handle various real-world applications involving changing deployment environments, \emph{e.g.} autonomous driving and intelligent surveillance.

Following the theoretical guarantee that the target distribution can be effectively approximated by the weighted combination of multiple source distributions~\cite{msda_theory_nips08,msda_theory_nips18}, recent works~\cite{MDAN,DCTN,M3SDA,MDDA} attempted to tackle the classification-based MSDA problem through aligning the feature distributions between source and target domains (or across different source domains) and combining the predictions of several domain-specific classification models. The core idea of these methods is to approach the conditional distribution of semantic label on target domain (\emph{i.e.} $p_{\mathcal{T}}(y|x)$) with the mixture of the conditional distributions learned for multiple source domains. Specifically, given a sample from target domain, these methods first derive the probability of its corresponding label using the classifiers trained for each source domain and then combine all predictions via a weighted average. Although such scheme is effective on several benchmark data sets of MSDA, its expressivity is still limited for the lack of following two important model capabilities. 

1) \textbf{Joint modeling across different domains.} The existing methods typically learn the conditional distribution of label on each domain in an independent way, which only model the dependency of label prediction on the statistics specific to a single domain. As a matter of fact, the interdependency between the statistics of various domains can also benefit the inference of a sample's semantic label. For instance, according to the similarity of a category-specific statistic, the correlated categories of different domains can be linked to each other, such that the cross-domain dependencies between these correlated categories can derive more precise predictions (\emph{e.g.} if an image is to be classified as vehicle, it should possess sufficient similarities with the vehicles and other related categories on various domains). Therefore, it is desirable to devise a unified model which can effectively capture the joint dependencies between a query sample and all the source and target domains.

2) \textbf{Learnable domain combination.} In most existing works, the domain combination is commonly attained by the weighted average using hand-craft or model-induced weights. In these methods, after learning the classification model for each source domain, the inference on a sample from target domain is performed by combining the predictions of different models according to the similarity scores of various source-target domain pairs. Such combination scheme relies on the heuristics of domain relations and is not learnable along with the model. It is more favorable to learn the domain combination from the data, in which the combination component of the model is directly optimized according to the learning objective. In this way, the model can better represent the relations between different domains under the guidance of the data. 

We would like to point out that the \emph{graphical structure} among various domains is informative to address the problems above. Specifically, the scope of valid joint distributions can be explicitly specified by a graphical structure, and such structure also enables learnable message passing across different domains. Motivated by these facts, in this work, we explore two types of graphical models, \emph{i.e.} \textbf{C}onditional \textbf{R}andom \textbf{F}ield for MSDA (\emph{CRF-MSDA}) and \textbf{M}arkov \textbf{R}andom \textbf{F}ield for MSDA (\emph{MRF-MSDA})\footnote{Note that, the CRF-MSDA and MRF-MSDA models differ from the conventional CRFs and MRFs, which are parameterized by local potential functions on subgraphs. Our models are instead parameterized by highly expressive deep neural networks. However, they share the similar working mechanism with these conventional methods on both graphical and probabilistic modeling, and are thus named after CRF and MRF.}. For joint modeling across various domains, both models introduce an additional set of random variables, named as prototypes~\cite{prototype,semantic,GraphLoG}, which serve as the representative embeddings of the semantic categories on all domains. On such basis, these two models learn two kinds of distributions over query sample and prototypes, in which the domain combination is intrinsically included and is thus learnable along with the whole model. These two graphical models are defined as follows.

\textbf{CRF-MSDA} seeks to model the conditional distribution of label for a query sample and all prototypes simultaneously. In specific, we first construct a graph over the query sample and the prototypes of different domains, in which the connection weight between two nodes is determined by the similarity of their features. We then employ a graph neural network (GNN) to propagate the local messages on the graph and use a linear classifier to predict the label of each node. During the learning phase, a global constraint is employed for the category-level alignment between different domains, 
and a local constraint is applied to promote the feature compactness surrounding the prototypes. In this model, the domain combination is achieved by the message passing between the prototypes of various domains, and such combination can be learned along with the GNN. 

\textbf{MRF-MSDA} aims to model the joint distribution of a query sample and all prototypes conditioned on a Markov network over them. For the MSDA problem, we consider a positive Markov network where all the prototypes belonging to the same category are connected, and the query sample is linked to the prototype associated to its corresponding domain and category. Also, some negative networks are derived by modifying the edges of the positive one. We optimize the joint distributions specified by various Markov networks through contrasting all the positive networks in a mini-batch with all negative ones.
In this way, the embedding of query sample is encouraged to be similar with the prototypes of its corresponding category and dissimilar with the ones of other categories. On such basis, we derive the classification probability for a query sample by summing the joint likelihoods over several specific Markov networks which link the query sample to the prototypes within the same category but from different domains. 
Such scheme attains domain combination, and it can be learned with the supervision from ground-truth labels. Compared to CRF-MSDA, the learning of MRF-MSDA involves multiple Markov networks (\emph{i.e.} positive and negative ones) for a single query sample, and thus more relational patterns between the query sample and prototypes can be learned. This property endows MRF-MSDA with stronger model expressivity. 

Compared to the conference paper~\cite{ltc_msda}, this journal work makes the following additional contributions:
\begin{itemize}
    \item We explicitly point out two important capabilities of an MSDA model, \emph{i.e.} the joint modeling across different domains and the learnable domain combination.
    \item We re-organize the LtC-MSDA approach proposed in the conference paper under the framework of CRF, deriving the CRF-MSDA model. 
    \item We novelly design a model that fully owns the two capabilities above. This model is designed based on the philosophy of MRF, called MRF-MSDA. Compared with CRF-MSDA that only models the dependency among labels, MRF-MSDA can jointly capture the dependency among observations and labels.
    \item We experimentally verify the superior performance of MRF-MSDA over CRF-MSDA, and MRF-MSDA establishes a new state-of-the-art on multiple MSDA benchmarks. 
\end{itemize}

\section{Related Work} \label{sec2}

\par{\textbf{Unsupervised Domain Adaptation (UDA).}}
UDA aims to generalize a model learned from a labeled source domain to another target domain without labels. Some previous methods attempts to narrow the domain shift between source and target domains via minimizing an explicit domain discrepancy metric, \emph{e.g.} Maximum Mean Discrepancy (MMD) \cite{two-sample_test,domain_confusion}, Weighted MMD \cite{w-mmd}, Multi-Kernel MMD \cite{dan, dan_pami} and Wasserstein Distance \cite{w-distance,sliced_w-distance,nw}. Also, aligning the second-order statistics is explored in \cite{deepcoral} to restrict the domain-invariance between two domains. 
Another group of methods perform adaptation by employing adversarial learning to align the source and target domains. Among these approaches, a domain discriminator is introduced to encourage domain-invariant features \cite{revgrad,adda,cdan, alda, gradually_vr,dada, Xu_2021_ICCV}. On par with the feature-level adaptation, generative models conduct distribution alignment on pixel level by image translation \cite{pixel-level,couple_gan}, style transfer \cite{style_transfer} or image generation \cite{reconstruction-classification,gta,domain_mixup,bi_gen}.
Cycle-consistency is also constrained to enforce the consistency of relevant semantics during distribution alignment \cite{cycada,joint_pixel_feature, ps_vae}.
Recently, a group of approaches performs category-level domain adaptation through utilizing dual classifier \cite{max_discrepancy,sliced_w-distance,fada}, domain prototype \cite{semantic,transferrable_proto,cross_domain_detection} or pseudo labels of target data \cite{collaborative,progressive,minimax_entropy, implicit_alignment}. 
There are also other domain adaptation approaches that focus on designing model components for domain transfer~\cite{transnorm} and exploring the transferability of label predictions~\cite{larger_norm,bnm,mcc}. 

To better exploit the structural dependency between the samples/categories of source and target domain, some existing methods~\cite{gcan,progressive_graph,heterogeneous} propose to construct relational graphs between two domains and perform domain alignment upon such inter-domain graphs. However, all these methods aim to tackle the UDA problem and cannot be trivially transferred to the setting with multiple source domains. By comparison, our work studies the graphical models upon multiple source domains and a target domain.


\par{\textbf{Multi-Source Domain Adaptation (MSDA).}}
In comparison with the conventional single-source domain adaptation, MSDA assumes data are collected from multiple source domains with different distributions, which is a more practical and difficult scenario. 
Early theoretical analysis \cite{msda_theory_nips08,bounds_DA} gave strong guarantees for representing target distribution as the weighted combination of source distributions to address the MSDA problem. 
Based on these works, Hoffman \emph{et al.} \cite{msda_theory_nips18} derived normalized solutions to determine the distribution-weighted combination. Recently, Zhao \emph{et al.} \cite{MDAN} proposed to align target domain to multiple source domains globally by adversarial learning. Xu \emph{et al.} \cite{DCTN} deployed multi-way adversarial learning and combined source-specific perplexity scores for target predictions. Peng \emph{et al.} \cite{M3SDA} introduced the idea of matching the high-order moments between domain-specific feature representations. In \cite{MDDA}, source distilling mechanism is designed to fine-tune the separately pre-trained feature extractor and classifier. 
CMSS~\cite{cmss} designed a dynamic curriculum to iteratively select the best source samples for aligning to the target. Li \emph{et al.}~\cite{meta_msda} enhanced model's domain adaptation performance by meta-learning.


\emph{Improvements over existing methods.} Previous works~\cite{MDAN,DCTN,M3SDA,MDDA} typically model the conditional distribution of semantic label on each domain in an independent way, and the label predictions from these domain-specific models are further combined to approach the conditional distribution on target domain. In contrast, in this work, we explore the joint modeling across all source and target domains. The domain combination is intrinsically contained in the proposed CRF-MSDA/MRF-MSDA model and thus can be learned in a joint fashion. 

\par{\textbf{Conditional Random Field (CRF) for Vision.}} 
CRFs are a class of probabilistic graphical modeling methods in which a set of observed variables $X$ and another set of unobserved ones $Y$ are considered, and these methods aim to model the conditional distribution $p(Y|X)$ utilizing the structure information among different variables. 
The concept of CRF was first proposed by Lafferty \emph{et al.}~\cite{crf} and applied to the field of segmenting and labeling text sequences, in which the label prediction on each observation well depends on the results of previous steps. 
Because of the strong capability of learning and inference on structured data, CRF-based approaches have been widely explored on various computer vision problems involved structured prediction, 
\emph{e.g.} segmentation~\cite{crf_seg_vem, crf_seg_yuan, crf_seg_3d}, image denoising~\cite{crf_denoise_roth, crf_denoise_vemu}, stereo reconstruction~\cite{crf_stereo_boykov, crf_stereo_xue} and super-resolution~\cite{crf_sr_tappen, crf_sr_wang}. 
These approaches mainly utilize the interrelationships among adjacent pixels/super-pixels. By comparison, for the MSDA task, we focus on the interdependencies among the semantic categories on different domains, and a CRF-MSDA model is proposed to perform structured prediction. In addition, compared to previous works, the CRF model established in this work is defined on the latent space instead of upon input images. 

\par{\textbf{Markov Random Field (MRF) for Vision.}} 
MRF is a probabilistic graphical model for joint distribution modeling over a set of random variables, which defines a family of joint distributions that can be factorized upon an undirected graph. 
MRFs were first introduced into the vision field by the work of Geman and Geman~\cite{mrf}, and their proposed MRF framework can express a wide variety of spatially varying priors, which is proved to benefit the image restoration task. Due to the effectiveness on capturing the interdependencies existing in different components of the data, 
MRF-based methods have been successfully adapted to many computer vision problems such as image restoration~\cite{mrf_restore_sun, mrf_restore_zhang}, segmentation~\cite{mrf_seg_held, mrf_seg_liu, mrf_seg_bao}, texture analysis~\cite{mrf_texture_cross} and optical flow prediction~\cite{mrf_flow_heitz, mrf_flow_lem}, in which obvious performance gain has been observed. In these works, a fixed Markov network is commonly adopted to model the joint distribution of different variables. By comparison, in the proposed MRF-MSDA method, we consider multiple Markov networks for each set of observations, so that more relational patterns can be explored. Furthermore, MRF is generally employed as a generative model for approximating the data distribution, while our MRF-MSDA model can be naturally used for discriminative modeling by summing the joint likelihoods over several specific Markov networks.


\begin{figure*}[t]
	\centering
	\includegraphics[width=0.95\textwidth]{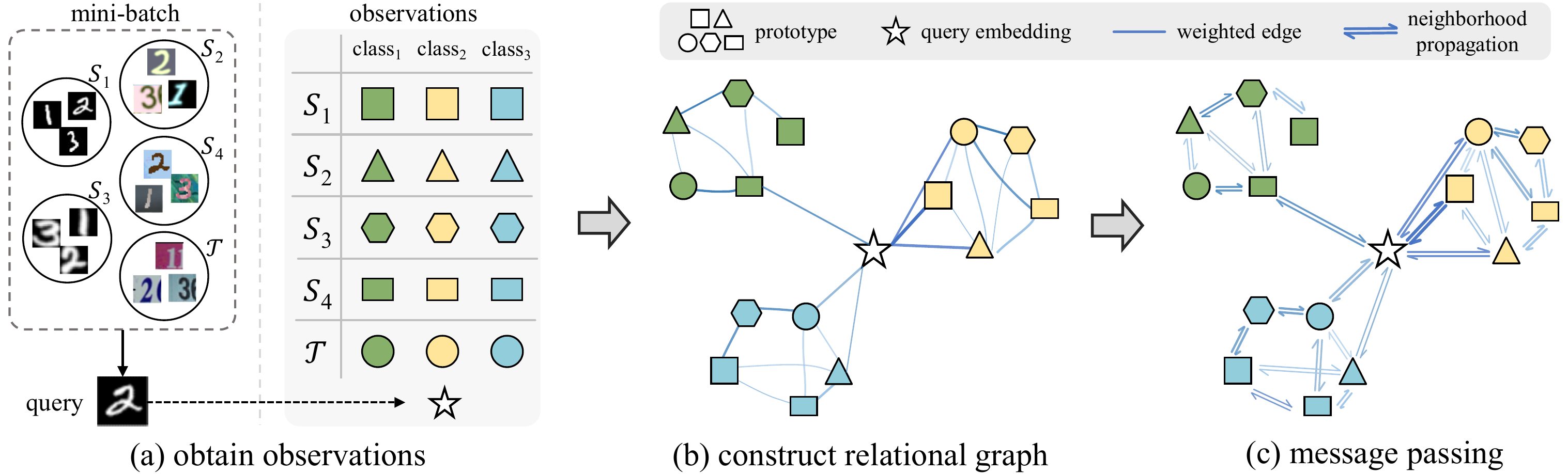}
	\vspace{-0.3cm}
	\caption{\textbf{Illustration of CRF-MSDA.} (a) Given a mini-batch of samples from various domains, a query sample from the mini-batch together with all prototypes serve as the observations. (b) A relational graph is constructed over the observations. Note that, this graph should be fully-connected, while we omit some edges in it for better visualization.} (c) The labels of all observations are predicted via the local message passing on the graph. 
	\label{fig_crf}
	\vspace{-0.3cm}
\end{figure*}


\section{Conditional Random Field for \\ Multi-Source Domain Adaptation} \label{sec3}

\subsection{Problem Definition} \label{sec3_1}

In Multi-Source Domain Adaptation (MSDA), there are $M$ source domains $\mathcal{S}_1$, $\mathcal{S}_2$, $\cdots$, $\mathcal{S}_M$. The source domain $S_m = \{(x^{\mathcal{S}_m}_i, y^{\mathcal{S}_m}_i)\}^{N_{\mathcal{S}_m}}_{i=1}$ contains $N_{\mathcal{S}_m}$ \emph{i.i.d.} labeled samples, where $x^{\mathcal{S}_m}_i$ follows the source distribution $p_{\mathcal{S}_m}(x)$ and $y^{\mathcal{S}_m}_i \in \{1, 2, \cdots, K\}$ ($K$ is the number of categories) denotes its corresponding label. Similarly, the target domain $\mathcal{T} = \{x^\mathcal{T}_j\}^{N_\mathcal{T}}_{j=1}$ is represented by $N_\mathcal{T}$ \emph{i.i.d.} unlabeled samples, where $x^\mathcal{T}_j$ follows the target distribution $p_{\mathcal{T}}(x)$. In addition, on all the source and target domains, we define a prototype (\emph{i.e.} a representative feature embedding) for each category, denoted as $\mathbb{C} =  \{ \{ c^{m}_{k} \}_{k=1}^{K} \}_{m=1}^{M+1}$, where target domain is regarded as the $(M+1)$-th domain in this notation.  

Given a query sample $q$ from an arbitrary domain, the Conditional Random Field for MSDA (\emph{CRF-MSDA}) considers an observed variable set $X$ and an output variable set $Y$. The query sample's embedding $z_q$ and all prototypes are deemed as observed variables, \emph{i.e.} $X = \{ z_q, c^{1}_{1}, \cdots, c^{M+1}_{K} \}$, and the semantic labels of these observations serve as the outputs, \emph{i.e.} $Y = \{ y_q, y^{1}_{1}, \cdots, y^{M+1}_{K} \}$. CRF-MSDA aims to model the conditional distribution $p(Y|X)$, in which a graph is constructed over the observed variables and their labels are predicted based on the local message passing on the graph. A graphical illustration of CRF-MSDA is shown in Fig.~\ref{fig_crf}. Next, we introduce the detailed learning and inference scheme of the CRF-MSDA approach. 


\subsection{Model Learning} \label{sec3_2}

The CRF-MSDA model seeks to learn the conditional distribution of labels for the observed variables defined above. Specifically, for each learning step, a mini-batch of query samples from various domains are given, and these samples are mapped to the latent space by a feature extractor to update prototypes. After that, we structure each query sample and all prototypes as a graph, and a GNN is employed to perform local message propagation on this graph, which derives the feature representations combining the information from different domains for the observations. Upon these representations, a linear classifier predicts the categorical probability for each observed variable, and the ground-truth labels are used for supervision. In addition, we further introduce a global and a local constraint for domain alignment and feature compactness, respectively. The details are presented in the following parts.


\subsubsection{Prototype Maintenance} \label{sec3_2_1}

During the learning phase, the prototypes are updated by the sampled mini-batches to better represent the data. Specifically, for each learning step, we sample a mini-batch $B$ constituted by sets of query samples from all the source and target domains, \emph{i.e.} $B = \{ \widehat{\mathcal{S}}_1, \widehat{\mathcal{S}}_2, \cdots , \widehat{\mathcal{S}}_M, \widehat{\mathcal{T}} \}$, and the estimations of prototypes are derived on this mini-batch. For the source domain $\mathcal{S}_m$ ($1 \leqslant m \leqslant M$), the estimated prototype $\widehat{c}^{m}_{k}$ is defined as the mean embedding of all samples belonging to class $k$ in the query sample set $\widehat{\mathcal{S}}_m$:
\begin{equation} \label{eq1}
\widehat{c}^{m}_{k} = \frac{1}{| \widehat{\mathcal{S}}^k_{m} |} \sum_{(x^{\mathcal{S}_m}_i, y^{\mathcal{S}_m}_i) \in \widehat{\mathcal{S}}^k_{m}} f(x^{\mathcal{S}_m}_i) ,
\end{equation}
where $\widehat{\mathcal{S}}^k_m$ is the set of all samples belonging to class $k$ in $\widehat{\mathcal{S}}_m$, and $f$ stands for the feature extractor which maps an image to a low-dimensional embedding vector. 

For the target domain $\mathcal{T}$, since the ground-truth label is unavailable, we first assign pseudo labels for the samples in $\widehat{\mathcal{T}}$ via the pseudo labeling strategy proposed by \cite{collaborative}, and the estimated prototype $\widehat{c}^{M+1}_{k}$ for class $k$ on target domain is defined as below:
\begin{equation} \label{eq2}
\widehat{c}^{M+1}_{k} = \frac{1}{| \widehat{\mathcal{T}}_k |} \sum_{(x^{\mathcal{T}}_i, \widehat{y}^{\mathcal{T}}_i) \in \widehat{\mathcal{T}}_k} f(x^{\mathcal{T}}_i) ,
\end{equation}
where $\widehat{y}^{\mathcal{T}}_i$ is the pseudo label assigned to $x^{\mathcal{T}}_i$, and $\widehat{\mathcal{T}}_k$ denotes the set of all samples labeled as the $k$-th category in $\widehat{\mathcal{T}}$. 

Using these mini-batch-induced estimations, we update the prototypes on various domains through an exponential moving average scheme: 
\begin{equation} \label{eq3}
c^{m}_{k} \leftarrow \beta c^{m}_{k} + (1 - \beta) \widehat{c}^{m}_{k}, \quad \ m = 1, 2, \cdots , M + 1 ,
\end{equation}
where $\beta$ denotes the exponential decay rate, and it is fixed as $0.7$ in all experiments. Such maintenance strategy can suppress the variance introduced by mini-batch sampling and derive smoother prototype estimations. In the literature~\cite{adam,semantic,moco}, similar strategies have been explored to stabilize the learning process via smoother global variables. 


\subsubsection{Graphical Modeling} \label{sec3_2_2}

In the CRF-MSDA model, we predict the labels of observed variables under the context determined by a graph, which models the conditional distribution $p(Y|X)$. In specific, for a query sample $q \in B$, we define the observed variable set with its embedding $z_q = f(q)$ and all prototypes, \emph{i.e.} $X = \{ z_q, c^{1}_{1}, \cdots, c^{M+1}_{K} \}$, and these observations are further structured as a graph $\mathcal{G} = ( \mathcal{V}, \mathcal{E} )$. In this graph, the node set $\mathcal{V}$ is identical to $X$ in which all nodes are represented by the embedding vectors with the same dimension, and the edge set $\mathcal{E} = \{ (u, v, A_{uv}) \}$ describes the relations among observations, where $A_{uv}$ denotes the adjacency weight between node $u$ and $v$. In practice, we derive the adjacency weight $A_{uv}$ by applying a radial basis function (RBF) kernel $\mathcal{K}$ upon the embeddings of two nodes:
\begin{equation} \label{eq4}
A_{uv} = \mathcal{K}(X_u, X_v) = \textrm{exp} \Big ( - \frac{|| X_u - X_v ||^2_2}{2 \sigma^2} \Big ) ,
\end{equation}
where $X_u$ and $X_v$ stand for the embedding of node $u$ and $v$, and $\sigma$ is the bandwidth parameter. Note that, the adjacency weights between all node pairs form the adjacency matrix of the graph, \emph{i.e.} $\mathbf{A} \in \mathbb{R}^{|\mathcal{V}| \times |\mathcal{V}|}$. 

Based on such a graph, we seek to learn effective node representations aggregating the information from neighbors and perform label prediction in a factorized way:
\begin{equation} \label{eq5}
p(Y|X) = \prod_{v \in \mathcal{V}} p(y_v | X) .
\end{equation}
Following the above formulation, a Graph Neural Network (GNN) $g$ is employed to produce node representations by propagating messages among different nodes, and, over these representations, a linear classifier $c$ outputs the classification probability for each node. In specific, the label of node $v$ is predicted as below:
\begin{equation} \label{eq6}
\mathbf{H} = g(\mathcal{G}, \mathbf{A}), \quad \hat{y}_{v} = p(y_v | X) = c(h_v) ,
\end{equation}
where $\mathbf{H} \in \mathbb{R}^{|\mathcal{V}| \times d}$ are the representations of all nodes produced by GNN ($d$ indicates the dimensionality), $h_v$ is the representation of node $v$, and $\hat{y}_{v}$ denotes the label prediction for that node. 


\subsubsection{Learning Objectives} \label{sec3_2_3}

For model learning, we aim to promote the discriminability and domain-invariance of feature representations, and these two objectives are pursued by constraining two kinds of objective functions for classification and alignment, respectively. The details are stated as follows.

\textbf{Classification constraints.} We define several classification constraints over the label predictions to enhance features' discriminability. In the observation set $X$, the prototypes are labeled by their corresponding category (\emph{e.g.} the prototype $c^m_k$ belongs to class $k$), which defines the following cross-entropy objective function:
\begin{equation} \label{eq7}
\mathcal{L}^{proto}_{cls} = - \frac{1}{(M+1)K} \sum_{m=1}^{M+1} \sum_{k=1}^{K} \, \log \big( \hat{y}^m_{k} [k] \big) ,
\end{equation}
where $\hat{y}^m_{k} [k]$ denotes the classification probability of prototype $c^m_k$ for the $k$-th category, and this class prediction is performed upon the post-GNN representation of prototype $c^m_k$, which better represents its corresponding semantic category via message passing. For the query sample $q$, when it is from source domains, the ground-truth label $y_q$ is available. Using all the source domain samples in mini-batch $B$ as query, we derive the supervised objective function for source domain as below:
\begin{equation} \label{eq8}
\mathcal{L}^{src}_{cls} = - \frac{1}{M} \sum_{m=1}^M \, \Big ( \mathbb{E}_{(q, y_q) \in \widehat{\mathcal{S}}_{m}} \log \big( \hat{y}_{q} [y_q] \big) \Big) ,
\end{equation}
where $\hat{y}_{q} [y_q]$ stands for the query sample's classification probability for the category specified by its label. In another case, when the query sample is drawn from the target domain, we cannot access the ground-truth annotation. Therefore, we resort to an entropy-induced constraint which is able to facilitate more deterministic predictions on the samples from target domain:
\begin{equation} \label{eq9}
\mathcal{L}^{tgt}_{cls} = - \mathbb{E}_{q \in \widehat{\mathcal{T}}} \sum_{k=1}^K  \, \hat{y}_{q} [k] \, \mathrm{log} \, \big( \hat{y}_{q} [k] \big) .
\end{equation}

For correctly classifying the nodes of various graphs established with different query samples from the mini-batch, the overall classification objective function is composed of three terms for prototypes, source domain queries and target domain queries, respectively:
\begin{equation} \label{eq10}
\mathcal{L}_{cls} = \mathcal{L}^{proto}_{cls} + \mathcal{L}^{src}_{cls} + \mathcal{L}^{tgt}_{cls} .
\end{equation}


\textbf{Alignment constraints.} Besides pursuing feature discriminability, we also expect the feature distributions of various domains to be invariant, and, especially, such domain-invariance is better to be attained on category-level. Formally, the marginal distribution of the samples from the source and target domain can be expressed as the summation over the conditional distributions associated to different categories:
\begin{equation} \label{eq11}
p_{\mathcal{S}_m}(x) = \sum_{y \in \mathbb{Y}} \, p_{\mathcal{S}_m}(y) \, p_{\mathcal{S}_m}(x | y) ,
\end{equation}
\begin{equation} \label{eq12}
p_{\mathcal{T}}(x) = \sum_{y \in \mathbb{Y}} \, p_{\mathcal{T}}(y) \, p_{\mathcal{T}}(x | y) ,
\end{equation}
where $\mathbb{Y}$ denotes the set of all categories. Under the assumption that the marginal distribution of category $p(y)$ is identical across different domains (\emph{i.e.} the proportions of the samples from various categories are domain-invariant), the goal is to align each conditional distribution $p(x|y)$ ($y \in \mathbb{Y}$) over all domains. To realize such a goal, we pursue the category-level domain alignment on the global level of the latent space, and the feature compactness surrounding various prototypes is constrained from a local point of view.

For the global objective, we expect the relevance between two arbitrary categories to be consistent on all domains. Specifically, we extract the first $(M+1)K$ rows and columns of the adjacency matrix, denoted as $\widetilde{\mathbf{A}} = \mathbf{A}^{1:(M+1)K}_{1:(M+1)K}$, where the block matrix $\widetilde{\mathbf{A}}_{i,j} = \widetilde{\mathbf{A}}^{(i-1)K+1:iK}_{(j-1)K+1:jK}$ ($1 \leqslant i,j \leqslant M+1$) measures all categories' relevance between the $i$-th and $j$-th domain. When various domains are well aligned on category level, these block matrices should be similar to each other, which leads to the following objective function for domain alignment:
\begin{equation} \label{eq13}
\mathcal{L}_{global} = \frac{1}{(M+1)^4} \sum_{i,j,m,n=1}^{M+1} || \widetilde{\mathbf{A}}_{i,j} - \widetilde{\mathbf{A}}_{m,n} ||_{F} ,
\end{equation}
where $|| \cdot ||_{F}$ is the Frobenius norm. In this function, the intra-class invariance is boosted by the constraints on block matrices' main diagonal elements, and the consistency of inter-class relationships is promoted by the constraints on other elements of block matrices. 

For the local objective, we expect the query samples to be compactly embedded around their corresponding prototypes, which eases the category-level alignment by deriving more separated features among distinct categories. In specific, we constrain the embeddings of the samples in mini-batch $B$ with the following objective function for feature compactness:
\begin{equation} \label{eq14}
\begin{split}
\mathcal{L}_{local} = \frac{1}{| B |} \, \sum_{k=1}^{K} \Bigg ( \sum_{m=1}^{M}  & \, \sum_{(x^{\mathcal{S}_m}_i,  y^{\mathcal{S}_m}_i) \in \widehat{\mathcal{S}}^k_{m}} || f(x^{\mathcal{S}_m}_i) - c^{m}_{k} ||^2_2  \\
& + \sum_{(x^{\mathcal{T}}_i, \widehat{y}^{\mathcal{T}}_i) \in \widehat{\mathcal{T}}_k} || f(x^{\mathcal{T}}_i) - c^{M+1}_{k} ||^2_2 \Bigg ) ,
\end{split}
\end{equation}
where $\widehat{\mathcal{S}}^k_{m}$ ($1 \leqslant m \leqslant M$) and $\widehat{\mathcal{T}}_k$ represent the samples belonging to class $k$ in the sample set $\widehat{\mathcal{S}}_{m}$ and $\widehat{\mathcal{T}}$, respectively. 


\begin{figure*}[t]
	\centering
	\includegraphics[width=0.95\textwidth]{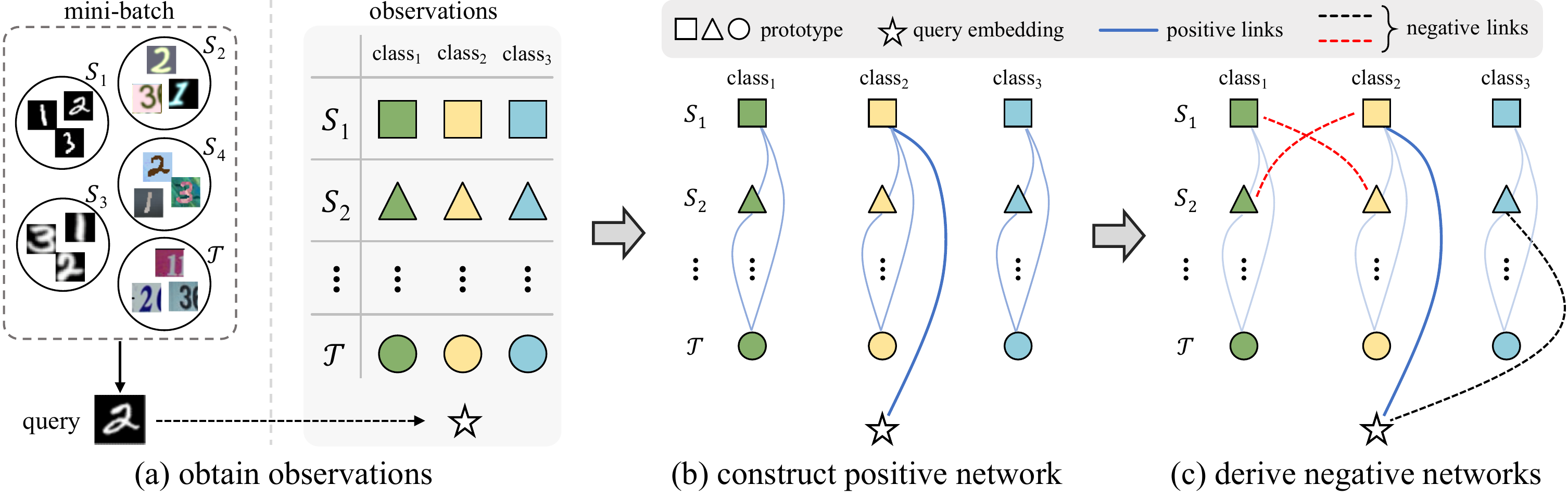}
	\vspace{-0.3cm}
	\caption{\textbf{Illustration of MRF-MSDA.} (a) A query sample from the mini-batch and all prototypes serve as the observations. (b) A positive Markov network is constructed to connect the prototypes within the same category and the corresponding query-prototype pair. (c) Negative Markov networks are derived by randomly modifying some edges in the positive network. The model learns the correct connection through contrasting the positive network with negative ones, which enables better label prediction.} 
	\vspace{-0.3cm}
	\label{fig_mrf}
\end{figure*}


\textbf{Overall learning objective.} Combining the classification and alignment constraints, the overall learning objective with respect to feature extractor $f$, GNN $g$ and classifier $c$ is defined as below: 
\begin{equation} \label{eq15}
\min \limits_{f, g, c} \, \mathcal{L}_{cls} + \lambda_1 \mathcal{L}_{global} + \lambda_2 \mathcal{L}_{local} ,
\end{equation}
where $\lambda_1$ and $\lambda_2$ are the trade-off parameters balancing among different learning objectives.


\subsection{Model Inference} \label{sec3_3}

After the learning phase, we store the feature extractor $f$, GNN model $g$, linear classifier $c$ and all prototypes $\mathbb{C} =  \{ \{ c^{m}_{k} \}_{k=1}^{K} \}_{m=1}^{M+1}$. In the inference phase, given a query sample $q$, we first extract its embedding $z_q$ with the extractor $f$ and combine the embedding with all prototypes to form the observation set $X = \{ z_q, c^{1}_{1}, \cdots, c^{M+1}_{K} \}$. After that, following the scheme in Sec.~\ref{sec3_2_2}, a graph $\mathcal{G}$ is constructed over the observations. Upon this graph, the GNN $g$ and linear classifier $c$ are consecutively applied to derive the label predictions for all nodes. Finally, we take the prediction for the node corresponding to the query sample as the output. 


\section{Markov Random Field for \\ Multi-Source Domain Adaptation} \label{sec4}

\subsection{Problem Definition} \label{sec4_1}

In this model, the definitions of source and target domains and prototypes follow those in Sec.~\ref{sec3_1}. Unlike the CRF-MSDA model, given a query sample $q$ from an arbitrary domain, the Markov Random Field for MSDA (\emph{MRF-MSDA}) seeks to model the joint distribution of all observed variables (\emph{i.e.} the query sample's embedding and all prototypes) conditioned on a Markov network $\mathcal{G}$, denoted as $p(X | \mathcal{G}) = p(z_q, c^{1}_{1}, \cdots, c^{M+1}_{K} | \mathcal{G})$.
Over all observations, a positive Markov network is formed to depict the desired interdependency among them. Specifically, all the prototypes belonging to the same category are connected, and the query sample is linked to the prototype associated to its corresponding domain and category. In addition, through modifying some edges in the positive network, the negative Markov networks are derived for comparison. Learning over these different networks guides the model to connect the query sample with correlated prototypes and thus enables label prediction. A graphical illustration of MRF-MSDA is presented in Fig.~\ref{fig_mrf}. The detailed learning and inference schemes are stated in the following sections. 


\subsection{Model Learning} \label{sec4_2}

Over a set of observations, the MRF-MSDA model is expected to be able to discriminate the positive Markov network from the negative ones through joint distribution modeling, and it can be further utilized for label prediction by summing the joint likelihoods over several specific Markov networks that link the query sample to the prototypes within a category.
Specifically, we represent the joint distribution of observations on a specific Markov network with an energy-based formulation, and the joint distributions for various Markov networks are learned via Noise Contrastive Estimation (NCE)~\cite{NCE_1,NCE_2}. Furthermore, the ground-truth labels of query samples are employed to supervise the joint-likelihood-induced label predictions. Instead of being updated via moving average as in CRF-MSDA, the prototypes in MRF-MSDA serve as model parameters and are learned along with the whole model. Next, we elucidate the details of model learning.


\subsubsection{Graphical Modeling} \label{sec4_2_1}

\textbf{Joint distribution modeling.} In the MRF-MSDA model, the joint distributions of observations are modeled over various Markov networks.
Specifically, for a query sample $q$, its embedding $z_q = f(q)$ together with all prototypes serve as the observed variables, \emph{i.e.} $X = \{ z_q, c^{1}_{1}, \cdots, c^{M+1}_{K} \}$. Note that, MRF-MSDA model uses a CNN encoder $f$ to map the query sample $q$ to a lower-dimensional embedding $z_q$, while the prototypes in this model are represented by learnable embedding vectors $\{c^1_1, \cdots, c^{M+1}_{K}\}$ following conventional graph embedding methods~\cite{deepwalk,line,node2vec}. Over these observations, it is expected that the prototypes within a same category are interrelated, and the query sample is most relevant to the prototype associated to its corresponding domain and category, which defines a positive Markov network $\mathcal{G}^+ = (\mathcal{V}, \mathcal{E}^+)$. In this network, the node set $\mathcal{V}$ is identical to the observation set $X$ where the embeddings of all nodes are with the same dimension, and the edge set $\mathcal{E}^+ = \{ (u,v) \}$ reflects the desired relationships among observations as stated above. We graphically illustrate the structure of $\mathcal{G}^+$ for an arbitrary query in Fig.~\ref{fig_mrf}(b). Based on the positive network $\mathcal{G}^+$, we randomly modify some edges in it to further construct $N_{neg}$ negative Markov networks $\{ \mathcal{G}_n^- = (\mathcal{V}, \mathcal{E}_n^-) \}_{n=1}^{N_{neg}}$ (the details about the edge modification scheme are stated in Sec.~\ref{sec5_1}). Upon a specific Markov network $\mathcal{G}$, we use an energy-based formulation to define the joint likelihood of the observations as follows:
\begin{equation} \label{eq16}
p(X|\mathcal{G}) = \frac{1}{Z} \exp \big( \! - \! f_E (X, \mathcal{G}) \big) ,
\end{equation}
\begin{equation} \label{eq17}
f_E(X, \mathcal{G}) = \frac{1}{\, \tau \,} \sum_{(u,v) \in \mathcal{E}} || X_u - X_v ||^2_2 ,
\end{equation}
where $Z$ stands for the partition function, $\tau$ denotes the temperature parameter, and $X_u$ and $X_v$ represent the embeddings of node $u$ and $v$ (these two nodes are connected in network $\mathcal{G}$). The energy function $f_E$ sums up the energies on all edges of the network. Using such joint likelihood definition, we perform model learning based on maximum likelihood estimation (MLE), and the concrete learning objective is introduced in Sec.~\ref{sec4_2_2}.


\textbf{Joint-likelihood-induced label prediction.} Considering the semantics underlying the observed variables, we propose to derive the classification probability of query sample using the joint likelihoods defined over several specific Markov networks. For example, we consider the case that only the prototypes within the same category are interrelated, and the query sample $q$ belongs to class $k$ and is from the $m$-th domain ($1 \leqslant m \leqslant M+1$), where the target domain is regarded as the $(M+1)$-th domain. The Markov network corresponding to such case is denoted as $\mathcal{G}^m_k$, in which $K$ cliques are formed among the prototypes of $K$ categories (\emph{i.e.} all the prototypes within a category are connected to each other), and the query sample is linked to prototype $c^m_k$. Using the joint likelihood of observations over $\mathcal{G}^m_k$, we define the probability that query sample $q$ is from the $k$-th category of the $m$-th domain as follows:
\begin{equation} \label{eq18}
p(y_d = m, y = k | q) = \frac{1}{\mathcal{N}} \, p(X | \mathcal{G}^m_k) ,
\end{equation}
\begin{equation} \label{eq19}
\mathcal{N} = \sum_{m=1}^{M+1} \sum_{k=1}^{K} p(X | \mathcal{G}^m_k) ,
\end{equation}
where the random variable $y_d$ represents the domain label, and $\mathcal{N}$ is the normalizing constant. Through summing the probability $p(y_d = m, y = k | q)$ over all domains, we derive the classification probability of query $q$ on class $k$ as below: 
\begin{equation} \label{eq20}
\hat{y}_{q} [k] = p(y = k | q) = \sum_{m=1}^{M+1} p(y_d = m, y = k | q) .
\end{equation}


\subsubsection{Learning Objectives} \label{sec4_2_2}

For learning the MRF-MSDA model, we aim at boosting the model's discriminative capability for label prediction and also maximizing the likelihoods on positive Markov networks while minimizing those on negative ones. These two learning objectives are pursued by classification constraints and maximum likelihood estimation (MLE), respectively. Detailed approaches are introduced as follows.

\textbf{Classification constraints.} We utilize two classification constraints to enhance model's discriminability on both source and target domains. In specific, for each learning step, we draw a mini-batch of query samples from source and target domains, denoted as $B = \{ \widehat{\mathcal{S}}_1, \widehat{\mathcal{S}}_2, \cdots , \widehat{\mathcal{S}}_M, \widehat{\mathcal{T}} \}$. Considering the unavailability of the ground-truth labels on target domain, we follow the formulations in Eqs.~\ref{eq8} and \ref{eq9} to obtain a supervised constraint $\mathcal{L}^{src}_{cls}$ for source domain and a label-free constraint $\mathcal{L}^{tgt}_{cls}$ for target domain.
For the discriminative modeling on these two kinds of domains, the overall classification objective function combines two constraints as below:
\begin{equation} \label{eq21}
\mathcal{L}_{cls} = \mathcal{L}^{src}_{cls} + \mathcal{L}^{tgt}_{cls} .
\end{equation}


\textbf{Maximum likelihood estimation (MLE).} Except for the discriminative modeling, we also expect that the model is able to identify the correct interrelationships among observations. We pursue such goal through enhancing the likelihoods for positive Markov networks and diminishing those for negative networks. This scheme guides the model to assign higher likelihoods to the networks that connect the query sample with correlated prototypes, which can benefit label prediction.

However, it is hard to directly optimize with the joint likelihood defined in Eq.~\ref{eq16} due to the intractability of evaluating the partition function exactly. As a substitute, inspired by the idea of Noise Contrastive Estimation (NCE)~\cite{NCE_1,NCE_2}, we propose to optimize upon the unnormalized joint likelihood, \emph{i.e.} $\tilde{p}(X|\mathcal{G}) = \exp \big( \! - \! f_E (X, \mathcal{G}) \big)$, by contrasting the positive Markov network with the negative ones. In practice, for constructing positive networks for the query samples from target domain, we again adopt the pseudo labeling scheme proposed by \cite{collaborative} to assign pseudo labels to the samples in $\widehat{\mathcal{T}}$. Formally, we define the following MLE-based objective function:
\begin{equation} \label{eq22}
\mathcal{L}_{MLE} = - \frac{1}{|B|} \sum_{q \in B} \Big( \tilde{p}(X|\mathcal{G}^+) - \frac{1}{N_{neg}} \sum_{n=1}^{N_{neg}} \tilde{p}(X|\mathcal{G}^{-}_n) \Big) ,
\end{equation}
where $|B|$ denotes the batch size, and $\mathcal{G}^+$ and $\mathcal{G}^{-}_i$ ($1 \leqslant n \leqslant N_{neg}$) are the positive and negative Markov networks for the query $q$, respectively. The partition function naturally vanishes in this expression after contrasting the joint likelihood defined over the positive network with the ones associated to negative networks. 

By optimizing with such objective function, two desired properties can be attained: (1) the embedding of the query sample is encouraged to approach the prototypes of its corresponding domain and category; (2) the prototypes from different domains but within the same category are aligned in the latent space, \emph{i.e.} achieving domain invariance.


\textbf{Overall objective.} In the MRF-MSDA model, the prototypes $\mathbb{C} =  \{ \{ c^{m}_{k} \}_{k=1}^{K} \}_{m=1}^{M+1}$ are optimized along with the feature extractor $f$ to minimize the classification and MLE-based objective functions as below:
\begin{equation} \label{eq23}
\min \limits_{f, \mathbb{C}} \, \mathcal{L}_{cls} + \alpha \mathcal{L}_{MLE} ,
\end{equation}
where $\alpha$ is the trade-off weight for the MLE objective. 


\subsection{Model Inference} \label{sec4_3}

When model learning is finished, we save the feature extractor $f$ and all prototypes $\mathbb{C}$. During inference, given a query sample $q$, its embedding $z_q$ is extracted by the feature extractor, and the observation set $X = \{ z_q, c^{1}_{1}, \cdots, c^{M+1}_{K} \}$ is formed by $z_q$ and all prototypes. After that, following the label prediction scheme proposed in Sec.~\ref{sec4_2_1}, we derive the classification probability for the query sample by summing the joint likelihoods of the observations over several specific Markov networks. 


\begin{table*}[t]
	\begin{spacing}{1.0}
		\centering
		\caption{The training setups on four different data sets.} \label{exp_setup}
		\setlength{\tabcolsep}{2.7mm}
		\begin{threeparttable}
		\begin{tabular}{c|c|c|c|c|c|c|c|c}
			\toprule[1.0pt]
			
			\multirow{2}{*}{data set} & \multirow{2}{*}{\# domains} & \multirow{2}{*}{\# classes} & \multirow{2}{*}{\begin{tabular}[c]{@{}c@{}}image\\ size\end{tabular}} & \multirow{2}{*}{backbone} &
			\multirow{2}{*}{\begin{tabular}[c]{@{}c@{}}batch\\ size\tnote{*} \end{tabular}} & \multirow{2}{*}{\begin{tabular}[c]{@{}c@{}}learning\\ rate\end{tabular}} & \multirow{2}{*}{\begin{tabular}[c]{@{}c@{}}\# training\\  epochs\end{tabular}} & \multirow{2}{*}{\begin{tabular}[c]{@{}c@{}}feature \\ dimension\end{tabular}} \\
			&         &               &           &         & &   &  \\ 		
			\hline
			
			Digits-five & 5       & 10      & $32 \times 32$      & 3 conv-2 fc      & 128 & $2 \times 10^{-4}$  & 100            & 2048               \\
			Office-31\cite{office_31_dataset}  & 3       & 31      & $252 \times 252$    & AlexNet  & 16   & $5 \times 10^{-5}$ & 100            & 4096               \\
		    PACS\cite{pacs}  & 4       & 7     & $224 \times 224$    & ResNet-18 & 16  & $5 \times 10^{-5}$ & 100             & 512              \\
			DomainNet\cite{M3SDA}  & 6       & 345     & $224 \times 224$    & ResNet-101 & 16  & $5 \times 10^{-5}$ & 20             & 2048              \\

			\bottomrule[1.0pt]
		\end{tabular}
	\begin{tablenotes}
		\scriptsize
		\item[*] Batch size here denotes the number of examples sampled from one domain in each iteration.
	\end{tablenotes}
	\end{threeparttable}
	\end{spacing}
\end{table*}

\subsection{Comparison between CRF-MSDA and MRF-MSDA} \label{sec4_4}

In this section, we compare the proposed CRF-MSDA and MRF-MSDA model from two aspects, \emph{i.e.} the model expressivity and the computational complexity, to shed the light on the effectiveness of these two types of graphical models. 

\subsubsection{Model Expressivity} \label{sce4_4_1}

Given a set of observations $X = \{ z_q, c^{1}_{1}, \cdots, c^{M+1}_{K} \}$ composed of a query sample embedding and the prototypes on all domains, CRF-MSDA seeks to model the joint distribution of their corresponding labels $Y = \{ y_q, y^{1}_{1}, \cdots, y^{M+1}_{K} \}$ conditioned on the observations, \emph{i.e.} $p(Y|X)$. By comparison, MRF-MSDA aims to model the joint distribution of both observations and labels, \emph{i.e.} $p(X,Y)$. Therefore, compared to CRF-MSDA, MRF-MSDA can not only capture the dependency among labels but also capture the dependency among different observations. As a matter of fact, the dependency among observations is useful to predict more accurate labels. For example, it can constrain the label predictions of correlated/uncorrelated observations to be similar/dissimilar. Such an advantage endows MRF-MSDA with stronger model expressivity.



\subsubsection{Computational Complexity} \label{sec4_4_2}

We compare the computational complexity of two models on processing a single query sample step by step. For feature extraction and label prediction steps, the time complexity of learning and inference are identical for both models. However, for the graph construction step, the time complexity of two phases are different, and we thus discuss the complexity of learning and inference separately.

\textbf{Feature extraction.} Given a query sample, both models employ a feature extractor to obtain the query's embedding, which possesses identical computational cost. 

\textbf{Graph construction.} The computational complexity of this step differs between learning and inference. In the learning phase, since the prototypes are online updated, the graph construction involves the computation of all prototypes and the query sample. The relational graph $\mathcal{G}$ constructed in the CRF-MSDA model requires to compute the pair-wise adjacency weight between $(M+1)K$ prototypes and a query sample, and thus the time complexity equals to $\mathcal{O}(M^2 K^2)$. For the MRF-MSDA model, a set of Markov networks $\mathbb{G} = \{ \{ \mathcal{G}^m_k \}_{k=1}^{K} \}_{m=1}^{M+1}$ are established. These networks have the same connections among prototypes (\emph{i.e.} $K$ cliques for $K$ categories) and a different edge linking the query sample to $(M+1)K$ distinct prototypes. In order to derive the joint likelihoods of the observations over these networks, $\frac{M(M+1)}{2} K$ pair-wise energies among prototypes and $(M+1)K$ energies between the query and each prototype are computed, which owns a time complexity of $\mathcal{O}(M^2 K)$. Therefore, MRF-MSDA is less computationally expensive than CRF-MSDA in this step during learning.

In the inference phase, the prototypes are fixed, and thus the adjacency weights (for CRF-MSDA) and the energies (for MRF-MSDA) among prototypes can be pre-computed. Therefore, given a query sample, the rest of computation is only between the query sample and prototypes, which has a time complexity of $\mathcal{O}(MK)$ for both models. In this way, CRF-MSDA and MRF-MSDA own an identical computational cost for graph construction during inference.

\textbf{Label prediction.} The CRF-MSDA model performs message passing on the constructed graph via a GNN model and predicts query's label by a linear classifier. By comparison, the label prediction of MRF-MSDA only requires the basic arithmetic calculations upon the joint likelihoods, which is model-free and more efficient.

In summary, for both learning and inference, the MRF-MSDA model is more computationally efficient than the CRF-MSDA counterpart in terms of processing a single query sample. In Sec.~\ref{sec6_3}, we further conduct an empirical time complexity analysis to verify the points above.


\section{Experiments} \label{sec5}

In this section, we first describe the experimental settings and then compare the proposed models with existing methods on various benchmark data sets of MSDA to demonstrate their effectiveness.


\begin{table*}[t]
	\begin{spacing}{1.0}
		\centering
		\caption{Classification accuracy (mean $\pm$ std \%) of various methods on five MSDA tasks of \emph{Digits-five}.} \label{table_digit}
		\setlength{\tabcolsep}{4.5mm}
		\begin{tabular}{c|c|c|c|c|c|c|c}
			\toprule[1.0pt]
			\multirow{1}{*}{Standards} & Methods & $\rightarrow$ \textbf{mm} & $\rightarrow$ \textbf{mt} &$\rightarrow$ \textbf{up} & $\rightarrow$ \textbf{sv} & $\rightarrow$ \textbf{syn}  & Avg  \\
			\hline
			\hline
			
			\multirow{5}{*}{\begin{tabular}[c]{@{}c@{}}Single\\Best\end{tabular} } 
			& Source-only  & 59.2$\pm$0.6  & 97.2$\pm$0.6  & 84.7$\pm$0.8  &  77.7$\pm$0.8 & 85.2$\pm$0.6 & 80.8 \\ 
			&DAN~\cite{dan} & 63.8$\pm$0.7 & 96.3$\pm$0.5 & 94.2$\pm$0.9 & 62.5$\pm$0.7 & 85.4$\pm$0.8 & 80.4\\
			&CORAL~\cite{deepcoral} & 62.5$\pm$0.7 & 97.2$\pm$0.8 & 93.5$\pm$0.8 & 64.4$\pm$0.7 & 82.8$\pm$0.7 & 80.1 \\
			&DANN~\cite{dann} & 71.3$\pm$0.6 & 97.6$\pm$0.8 & 92.3$\pm$0.9 & 63.5$\pm$0.8 & 85.4$\pm$0.8 & 82.0\\
			&ADDA~\cite{adda} & 71.6$\pm$0.5 & 97.9$\pm$0.8 & 92.8$\pm$0.7 & 75.5$\pm$0.5 & 86.5$\pm$0.6 & 84.8\\
			
			\hline
			\multirow{6}{*}{ \begin{tabular}[c]{@{}c@{}}Source\\Combine\end{tabular} } 
			&Source-only & 63.4$\pm$0.7 & 90.5$\pm$0.8 & 88.7$\pm$0.9 & 63.5$\pm$0.9 & 82.4$\pm$0.6 &77.7\\	
			& DAN~\cite{dan}  &  67.9$\pm$0.8 & 97.5$\pm$0.6  & 93.5$\pm$0.8  &  67.8$\pm$0.6 & 86.9$\pm$0.5  & 82.7 \\
			& DANN~\cite{dann}  & 70.8$\pm$0.8  & 97.9$\pm$0.7  &  93.5$\pm$0.8 &  68.5$\pm$0.5 & 87.4$\pm$0.9  & 83.6 \\
			& JAN~\cite{JAN} & 65.9$\pm$0.7 & 97.2$\pm$0.7 & 95.4$\pm$0.8 & 75.3$\pm$0.7 & 86.6$\pm$0.6 &84.1  \\ 
			& ADDA~\cite{adda}  & 72.3$\pm$0.7  & 97.9$\pm$0.6  & 93.1$\pm$0.8  &  75.0$\pm$0.8 & 86.7$\pm$0.6  &  85.0 \\

			& MCD~\cite{max_discrepancy} & 72.5$\pm$0.7 & 96.2$\pm$0.8 & 95.3$\pm$0.7 & 78.9$\pm$0.8 & 87.5$\pm$0.7 & 86.1\\
			\hline
			
			\multirow{7}{*}{ \begin{tabular}[c]{@{}c@{}}Multi-\\Source\end{tabular} } 
			& MDAN~\cite{MDAN} &69.5$\pm$0.3& 98.0$\pm$0.9& 92.4$\pm$0.7& 69.2$\pm$0.6& 87.4$\pm$0.5 & 83.3 \\
			& DCTN~\cite{DCTN} & 70.5$\pm$1.2 & 96.2$\pm$0.8 & 92.8$\pm$0.3 & 77.6$\pm$0.4 & 86.8$\pm$0.8 & 84.8\\
			& $\rm M^{3}SDA$~\cite{M3SDA} & 72.8$\pm$1.1 & 98.4$\pm$0.7 & 96.1$\pm$0.8 & 81.3$\pm$0.9 & 89.6$\pm$0.6 &87.7\\	
			& MDDA~\cite{MDDA} &78.6$\pm$0.6& 98.8$\pm$0.4& 93.9$\pm$0.5& 79.3$\pm$0.8& 89.7$\pm$0.7 & 88.1 \\
			
			& CMSS~\cite{cmss} &75.3$\pm$0.6& 99.0$\pm$0.1& 97.7$\pm$0.1& \textbf{88.4}$\pm$0.5& 93.7$\pm$0.2 & 90.8 \\

			& \textbf{CRF-MSDA} & 85.6$\pm$0.8 & 99.0$\pm$0.4 & 98.3$\pm$0.4 & 83.2$\pm$0.6 & 93.0$\pm$0.5 & 91.8 \\	
			& \textbf{MRF-MSDA} & \textbf{90.7}$\pm$0.7 & \textbf{99.2}$\pm$0.2 & \textbf{98.5}$\pm$0.4 & 85.8$\pm$0.7 & \textbf{94.7}$\pm$0.5 & \textbf{93.7} \\	
			
			\bottomrule[1.0pt]
		\end{tabular}
	\end{spacing}
\end{table*}


\subsection{Experimental Setup} \label{sec5_1}

\textbf{Model details.} For the CRF-MSDA model, we adopt a two-layer GCN~\cite{gcn_model} model to propagate messages among the observations, and, for each node of the relational graph, a linear classifier maps its $d$-dimensional feature representation to the $K$-dimensional classification probability. 
For the MRF-MSDA model, we consider two ways of deriving negative Markov networks based on a positive network: (1) The link between the query sample and its corresponding prototype is deleted, and we connect the query sample with any one of the rest $K-1$ prototypes within the same domain but belonging to different categories, which defines $K-1$ negative networks; (2) We additionally select two random prototypes associated with distinct categories and connect them, which defines other $N_2$ negative networks. In total, for each query sample, we employ $N_{neg} = N_2 + K - 1$ negative networks to contrast with the positive one.


\textbf{Training details.} We list the basic training settings on four different data sets in Tab.~\ref{exp_setup}. The setup differences on these data sets are mainly due to the distinction of data complexity, which follows the common experimental setups in the literature~\cite{DCTN,M3SDA,MDAN,ltc_msda}. In all experiments, we adopt an Adam~\cite{adam} optimizer (weight decay: $5 \times 10^{-4}$) to train the model. For all the comparisons in this section, we use the following parameter settings for two proposed models: (1) For CRF-MSDA, the trade-off parameters $\lambda_{1}$ and $\lambda_{2}$ are set as 20 and 0.001 respectively, and the bandwidth parameter $\sigma$ is set as 0.005; (2) For MRF-MSDA, the trade-off weight $\alpha$ is set as $1.0$, the temperature parameter $\tau$ is set as 0.1, and the negative sampling size $N_{neg}$ is set as $K + 5$ (\emph{i.e.} $N_2 = 6$ negative networks per query are sampled by the second sampling way stated above). All these parameter setups are determined by the grid search on the source domains' validation sets of the $\rightarrow$ \textbf{mm} task (an MSDA task on Digits-five data set). For simplicity, we use ``$\rightarrow D$" to denote the task of transferring from other domains to domain $D$. 
Our approach is implemented with PyTorch \cite{pytorch}, and the source code will be released for reproducibility.


\textbf{Performance comparison.} We compare our approach with state-of-the-art methods to verify its effectiveness. For the sake of fair comparison, we introduce three standards. (1) \emph{Single Best}: We report the best performance of single-source domain adaptation algorithm among all the sources. (2) \emph{Source Combine}: All the source domain data are combined into a single source, and domain adaptation is performed in a traditional single-source manner. (3) \emph{Multi-Source}: The knowledge learned from multiple source domains are transferred to target domain. For the first two settings, previous single-source UDA methods, \emph{e.g.} DAN~\cite{dan}, JAN~\cite{JAN}, DANN~\cite{dann}, ADDA~\cite{adda}, MCD~\cite{max_discrepancy}, are introduced for comparison. For the \emph{Multi-Source} setting, we compare our approach with several existing MSDA algorithms, \emph{e.g.} MDAN~\cite{MDAN}, DCTN~\cite{DCTN}, $\rm M^{3}SDA$~\cite{M3SDA}, MDDA~\cite{MDDA}, and CMSS~\cite{cmss}. We report the performance of these methods on Digits-five and DomainNet from Peng \emph{et al.}~\cite{M3SDA}, on Office-31 from Zhao \emph{et al.}~\cite{MDDA} and on PACS from Yang \emph{et al.}~\cite{cmss}.


\begin{table}[t]
	\begin{spacing}{1.0}
		\centering
		\small
		\caption{Classification accuracy (\%) of various methods on three MSDA tasks of \emph{Office-31}.} \label{table_office}
		\setlength{\tabcolsep}{2.0mm}
		\begin{tabular}{c|c|c|c|c|c}
			\toprule[1.0pt]
			\multirow{1}{*}{Standards} & Methods & $\rightarrow$ D & $\rightarrow$ W & $\rightarrow$ A & Avg \\
			\hline
			\hline
		
		    \multirow{5}{*}{\begin{tabular}[c]{@{}c@{}}Single\\Best\end{tabular} } 
			& Source-only  & 99.0  & 95.3  & 50.2  &  81.5 \\
			& RevGrad~\cite{revgrad}  & 99.2  & 96.4  &  53.4 & 83.0 \\
			& DAN~\cite{dan}  & 99.0  & 96.0  & 54.0  & 83.0 \\
			& RTN~\cite{RTN}  & 99.6  & 96.8  & 51.0  & 82.5 \\
			& ADDA~\cite{adda}  & 99.4  & 95.3  & 54.6  & 83.1 \\
			\hline	
			
			\multirow{6}{*}{ \begin{tabular}[c]{@{}c@{}}Source\\Combine\end{tabular} } 
			&Source-only & 97.1  &  92.0 &  51.6 &  80.2\\
			&DAN~\cite{dan} & 98.8  &  96.2 &  54.9 &  83.3\\	
			&JAN~\cite{JAN} & 99.4  &  95.9 &  54.6 &  83.3 \\
			&DANN~\cite{dann} & 99.2  &  95.8 &  55.2 &  83.4\\
			&ADDA~\cite{adda} & 99.2  &  96.0 &  55.9 & 83.7\\
			&MCD~\cite{max_discrepancy} & 99.5  &  96.2 &  54.4 &  83.4\\
			
			\hline	
			\multirow{6}{*}{ \begin{tabular}[c]{@{}c@{}}Multi-\\Source\end{tabular} }  
			& MDAN~\cite{MDAN}  & 99.2  & 95.4  &  55.2 &  83.3  \\	
			&DCTN~\cite{DCTN} & 99.6 & 96.9 & 54.9 & 83.8 \\	
			&$\rm M^{3}SDA$~\cite{M3SDA}& 99.4 & 96.2 & 55.4 & 83.7 \\			
			& MDDA~\cite{MDDA}   & 99.2  & 97.1 &  56.2 &  84.2  \\	
			& \textbf{CRF-MSDA} & 99.6 & 97.2 & \textbf{56.9} & 84.6 \\
			& \textbf{MRF-MSDA} & \textbf{99.7} & \textbf{97.4} & \textbf{56.9} & \textbf{84.7} \\	
			
			\bottomrule[1.0pt]
		\end{tabular}
	\end{spacing}
\end{table}


\subsection{Experiments on Digits-five} \label{section5_2}

\textbf{Data set.} The Digits-five data set is composed of five digital image domains, including MNIST (\textbf{mt})~\cite{mnist}, MNIST-M (\textbf{mm})~\cite{dann}, SVHN (\textbf{sv})~\cite{svhn}, USPS (\textbf{up})~\cite{usps} and Synthetic Digits (\textbf{syn})~\cite{dann}. Each domain contains ten categories corresponding to the digits ranging from 0 to 9. 
Following the setting in DCTN~\cite{DCTN}, we sample 25000 images for training, 6000 images for validation and 9000 images for test on MNIST, MINST-M, SVHN and Synthetic Digits, and the entire USPS data set serves as a domain. The reported results are averaged over five independent runs under the same configuration. 

\textbf{Results.} In Tab.~\ref{table_digit}, we compare the proposed CRF-MSDA and MRF-MSDA models with other works. Source-only stands for the model trained with only source domain data, which serves as the baseline. Compared to the state-of-the-art CMSS~\cite{cmss} approach, CRF-MSDA achieves notable performance gain on the ``$\rightarrow$ \textbf{mm}'' task and surpasses it in terms of average accuracy over all tasks. The MRF-MSDA model performs best on four of five tasks and obtains a $12.1\%$ performance increase relative to previous methods. These promising results illustrate the effectiveness of cross-domain joint modeling and learnable domain combination which are first explored in our approaches. MRF-MSDA outperforms CRF-MSDA on all five tasks, which mainly owes to its exploration of more diverse relational patterns over the observations by using positive and negative Markov networks. 


\subsection{Experiments on Office-31} \label{section5_3}

\textbf{Data set.} Office-31~\cite{office_31_dataset} is a classical domain adaptation benchmark with 31 categories and 4652 images. It contains three domains, \emph{i.e.} Amazon (A), Webcam (W) and DSLR (D), and the data are collected from office environments. The data of Amazon are collected from amazon.com, while the data of Webcam and DSLR are captured by web camera and digital single-lens reflex camera under different conditions. There are 2,817, 795 and 498 images in A, W and D, respectively. Our methods are evaluated by five independent runs, and, following MDDA~\cite{MDDA}, we report the mean accuracy.

\textbf{Results.} Tab.~\ref{table_office} compares our methods with existing algorithms on three tasks. The MRF-MSDA model outperforms the state-of-the-art method, MDDA \cite{MDDA}, with 0.5\% in terms of average accuracy, and the CRF-MSDA model performs comparably with MRF-MSDA. 
On this data set, our approaches do not have obvious superiority, which probably ascribes to two reasons. (1) First, performance saturation occurs on ``$\rightarrow$ D'' and ``$\rightarrow$ W'' tasks, in which the Source-only model achieves performance higher than 95\%. (2) Second, the Webcam and DSLR domains are highly similar, which restricts the benefit brought by cross-domain joint modeling in our framework, especially in ``$\rightarrow$ A'' task.


\begin{table}[t]
	    \begin{spacing}{1.2}
		\centering
	    \footnotesize
		\caption{Classification accuracy (mean $\pm$ std \%) of various methods on four MSDA tasks of \emph{PACS}.} \label{table_pacs}
		\setlength{\tabcolsep}{1.5mm}
		\begin{tabular}{c|c|c|c|c|c}
			\toprule[1.0pt]
			Methods & $\rightarrow$ A & $\rightarrow$ C & $\rightarrow$ P & $\rightarrow$ S & Avg \\
			\hline
			Source-only  & 76.0$\pm$0.9 & 73.3$\pm$0.8 & 91.7$\pm$0.6 & 64.2$\pm$1.8  & 76.3 \\
			MDAN~\cite{MDAN}  & 79.1$\pm$0.4 & 76.0$\pm$0.7 & 91.4$\pm$0.9 & 72.0$\pm$0.8 & 79.6 \\
			DCTN~\cite{DCTN}  & 84.7$\pm$0.7 & 86.7$\pm$0.6 & 95.6$\pm$0.8 & 71.8$\pm$1.0 & 84.7 \\
		    $\rm M^{3}SDA$~\cite{M3SDA}  & 89.3$\pm$0.4 & 89.9$\pm$1.0 & 97.3$\pm$0.3 & 76.7$\pm$2.9 & 88.3 \\
		    MDDA~\cite{MDDA}  & 86.7$\pm$0.6 & 86.2$\pm$0.7 & 93.9$\pm$0.7 & 77.6$\pm$0.9 & 86.1 \\
		    Meta-MCD~\cite{meta_msda} & 87.4$\pm$0.7 & 86.2$\pm$0.9 & 97.1$\pm$0.5 & 78.3$\pm$0.8 & 87.2 \\
		    CMSS~\cite{cmss}  & 88.6$\pm$0.4 & 90.4$\pm$0.8 & 96.9$\pm$0.3 & 82.0$\pm$0.6 & 89.5 \\
			\textbf{CRF-MSDA} & 90.2$\pm$0.5 & 90.5$\pm$0.6 & 97.2$\pm$0.5 & 81.5$\pm$0.7 & 89.9 \\	
			\textbf{MRF-MSDA}  & \textbf{92.2}$\pm$0.4 & \textbf{93.3}$\pm$0.6 & \textbf{98.0}$\pm$0.3 & \textbf{86.7}$\pm$0.8 & \textbf{92.6} \\	
			\bottomrule[1.0pt]
		\end{tabular}
		\end{spacing}
\end{table}


\begin{table*}[t]
\begin{spacing}{1.05}
\centering
\caption{Classification accuracy (mean $\pm$ std \%) of various methods on six MSDA tasks of \emph{DomainNet}.} \label{table_domainnet}
\setlength{\tabcolsep}{3.2mm}
\begin{tabular}{c|c|c|c|c|c|c|c|c}
	\toprule[1.0pt]
	\multirow{1}{*}{Standards} & Methods & $\rightarrow$ clp & $\rightarrow$ inf &$\rightarrow$ pnt & $\rightarrow$ qdr & $\rightarrow$ rel & $\rightarrow$ skt  & Avg \\
	\hline
	\hline
	
	\multirow{6}{*}{\begin{tabular}[c]{@{}c@{}}Single\\Best\end{tabular} } 
	& Source-only & 39.6$\pm$0.6 & 8.2$\pm$0.8 & 33.9$\pm$0.6 & 11.8$\pm$0.7 & 41.6$\pm$0.8 & 23.1$\pm$0.7 & 26.4\\
	&DAN~\cite{dan} &  39.1$\pm$0.5 & 11.4$\pm$0.8 & 33.3$\pm$0.6 & 16.2$\pm$0.4 & 42.1$\pm$0.7  & 29.7$\pm$0.9 & 28.6\\
	&JAN~\cite{JAN} &  35.3$\pm$0.7 & 9.1$\pm$0.6 & 32.5$\pm$0.7 & 14.3$\pm$0.6& 43.1$\pm$0.8  & 25.7$\pm$0.6 & 26.7\\
	&DANN~\cite{dann} &  37.9$\pm$0.7 & 11.4$\pm$0.9 & 33.9$\pm$0.6 & 13.7$\pm$0.6& 41.5$\pm$0.7  & 28.6$\pm$0.6 & 27.8\\
	&ADDA~\cite{adda} &  39.5$\pm$0.8 & 14.5$\pm$0.7 & 29.1$\pm$0.8 & 14.9$\pm$0.5& 41.9$\pm$0.8  & 30.7$\pm$0.7 & 28.4\\
	&MCD~\cite{max_discrepancy} & 42.6$\pm$0.3 &  19.6$\pm$0.8  & 42.6$\pm$1.0 & 3.8$\pm$0.6 &50.5$\pm$0.4&33.8$\pm$0.9& 32.2 \\
	\hline
	
	\multirow{6}{*}{ \begin{tabular}[c]{@{}c@{}}Source\\Combine\end{tabular} } 
	&Source-only & 47.6$\pm$0.5  & 13.0$\pm$0.4 & 38.1$\pm$0.5   & 13.3$\pm$0.4 & 51.9$\pm$0.9 & 33.7$\pm$0.5 & 32.9\\
	&DAN~\cite{dan}& 45.4$\pm$0.5&	12.8$\pm$0.9&	36.2$\pm$0.6&	15.3$\pm$0.4&	48.6$\pm$0.7&	34.0$\pm$0.5&	32.1  \\
	&JAN~\cite{JAN}& 40.9$\pm$0.4&	11.1$\pm$0.6& 35.4$\pm$0.5&	12.1$\pm$0.7&	45.8$\pm$0.6&	32.3$\pm$0.6&	29.6  \\
	&DANN~\cite{dann}& 45.5$\pm$0.6& 13.1$\pm$0.7&	37.0$\pm$0.7&	13.2$\pm$0.8&	48.9$\pm$0.7&	31.8$\pm$0.6& 32.6  \\
	&ADDA~\cite{adda}& 47.5$\pm$0.8& 11.4$\pm$0.7&	36.7$\pm$0.5&	14.7$\pm$0.5&	49.1$\pm$0.8&	33.5$\pm$0.5& 32.2  \\
	&MCD~\cite{max_discrepancy}& 54.3$\pm$0.6&	22.1$\pm$0.7&	45.7$\pm$0.6&	7.6$\pm$0.5&	58.4$\pm$0.7&	43.5$\pm$0.6& 38.5  \\
	
	\hline	
	\multirow{8}{*}{ \begin{tabular}[c]{@{}c@{}}Multi-\\Source\end{tabular} } 
	& MDAN~\cite{MDAN} &52.4$\pm$0.6& 21.3$\pm$0.8& 46.9$\pm$0.4& 8.6$\pm$0.6& 54.9$\pm$0.6& 46.5$\pm$0.7& 38.4 \\	
	&DCTN~\cite{DCTN} &48.6$\pm$0.7 & 23.5$\pm$0.6  &48.8$\pm$0.6  &7.2$\pm$0.5& 53.5$\pm$0.6 & 47.3$\pm$0.5 & 38.2 \\
	&$\rm M^{3}SDA$~\cite{M3SDA} &58.6$\pm$0.5& 26.0$\pm$0.9& 52.3$\pm$0.6& 6.3$\pm$0.6& 62.7$\pm$0.5& 49.5$\pm$0.8& 42.6 \\
	
	& MDDA~\cite{MDDA} &59.4$\pm$0.6& 23.8$\pm$0.8& 53.2$\pm$0.6& 12.5$\pm$0.6& 61.8$\pm$0.5& 48.6$\pm$0.8& 43.2 \\
	
	& Meta-MCD~\cite{meta_msda} &62.8$\pm$0.2& 21.4$\pm$0.1& 50.5$\pm$0.1& 15.5$\pm$0.2& 64.6$\pm$0.2& 50.4$\pm$0.1& 44.2 \\
	& CMSS~\cite{cmss} &\textbf{64.2}$\pm$0.2& 28.0$\pm$0.2& 53.6$\pm$0.4& 16.0$\pm$0.1& 63.4$\pm$0.1& 53.8$\pm$0.4& 46.5 \\
			
	& \textbf{CRF-MSDA} & 63.1$\pm$0.4 & \textbf{28.7}$\pm$0.5 & 56.1$\pm$0.5 & 16.3$\pm$0.5 & 66.1$\pm$0.6 & 53.8$\pm$0.6 & 47.4 \\
	& \textbf{MRF-MSDA} & 63.9$\pm$0.3 & \textbf{28.7}$\pm$0.4 & \textbf{56.3}$\pm$0.4 & \textbf{16.8}$\pm$0.4 & \textbf{67.1}$\pm$0.6 & \textbf{54.3}$\pm$0.5 & \textbf{47.9} \\
	
	\bottomrule[1.0pt]
\end{tabular}
\end{spacing}
\vspace{-2mm}
\end{table*}


\subsection{Experiments on PACS} \label{sec5_4}

\textbf{Data set.} The PACS~\cite{pacs} data set includes 4 domains, \emph{i.e.} Photo (P), Art paintings (A), Cartoon (C) and Sketch (S). Each domain contains 7 categories, and significant domain shift (\emph{i.e.} distinct painting styles) exists between different domains. Following two previous works~\cite{meta_msda, cmss}, only the approaches under the \emph{Multi-Source} setting are employed for comparison. The mean and standard deviation of model performance over five independent runs are presented. 

\textbf{Results.} In Tab.~\ref{table_pacs}, we report the performance of various methods on four tasks. It can be observed that the proposed CRF-MSDA model performs comparably with the CMSS~\cite{cmss} approach. MRF-MSDA achieves the highest accuracy on all four tasks, and, especially, a $4.7\%$ performance gain is obtained on the ``$\rightarrow$ S'' task. The superior performance of MRF-MSDA can be mainly ascribed to its explorations of diverse intra- and inter-domain relations, which enables more precise label prediction when the distributional gap between different domains is large.


\subsection{Experiments on DomainNet} \label{section5_5}

\textbf{Data set.} DomainNet~\cite{M3SDA} is by far the largest and most difficult data set for MSDA. It consists of around 0.6 million images and 6 domains, \emph{i.e.} clipart (clp), infograph (inf), painting (pnt), quickdraw (qdr), real (rel) and sketch (skt). Each domain includes the same 345 categories of common objects. The reported model performance is averaged over five independent runs using the same setting.

\textbf{Results.} The results of various approaches on DomainNet are presented in Tab.~\ref{table_domainnet}. CRF-MSDA and MRF-MSDA perform comparably on this data set, and the latter achieves the best performance on five of six tasks. In particular, a 1.4\% performance increase on average accuracy is gained by MRF-MSDA. The major challenge of this data set is the great complexity of data distribution, which is caused by two factors: (1) Large distributional gaps exist among different domains, \emph{e.g.} from real images to sketches; (2) The numerous semantic categories within each domain lead to more complex single-domain data distribution. The CRF-MSDA model mitigates such dilemma by conducting category-level domain alignment and promoting feature compactness, while the MRF-MSDA model approaches such complex data distribution by modeling the joint distributions over various observations, which is a more direct scheme and performs better in practice. 


\section{Analysis} \label{sec6}

In this section, we provide more in-depth analysis of the proposed methods to verify the effectiveness of major model components, in which both quantitative and qualitative studies are conducted for validation. 


\subsection{Ablation Study} \label{sec6_1}


\begin{table}[t]
        \begin{spacing}{1.1}
		\centering
		\caption{The performance of CRF-MSDA under four model configurations on \emph{Digits-five}.} \label{table_ablation_crf}
	    \footnotesize
		\setlength{\tabcolsep}{1.6mm}
			\begin{tabular}{cc|ccccc|c}
			\toprule[1.0pt]

			 $\mathcal{L}_{global}$ & $\mathcal{L}_{local}$ & $\rightarrow$ \textbf{mm} & $\rightarrow$ \textbf{mt} &$\rightarrow$ \textbf{up} & $\rightarrow$ \textbf{sv} & $\rightarrow$ \textbf{syn}  & Avg \\ \specialrule{0em}{0pt}{0.8pt}
			
			\hline
			&  & 74.85 & 98.60 & 97.95 & 74.56 & 88.54 & 86.90 \\
			$\checkmark$ &   & 82.49 & 98.97 & 98.06 & 81.64 & 91.70 & 90.57  \\
			& $\checkmark$ & 79.57 & 98.64 & 98.06 & 78.66 & 90.16 & 89.02 \\
			$\checkmark$  & $\checkmark$ & \textbf{85.56} & \textbf{98.98} & \textbf{98.32} & \textbf{83.24} & \textbf{93.04} & \textbf{91.83} \\
			\bottomrule[1.0pt]		
			\end{tabular}
			\end{spacing}
\end{table}


\subsubsection{Ablation Study for CRF-MSDA} \label{sec6_1_1}

In this part, we analyze the effect of the global and local alignment objective functions on the CRF-MSDA model. In Tab.~\ref{table_ablation_crf}, we evaluate model's performance under four configurations on the Digits-five data set. In the baseline setting (\emph{1st} row), only the classification constraint (Eq.~\ref{eq10}) is utilized to optimize the model. On the basis of the baseline setting, the global alignment constraint $\mathcal{L}_{global}$ (Eq.~\ref{eq13}) can greatly enhance model's performance by performing category-level domain alignment (\emph{2nd} row). For the local alignment constraint $\mathcal{L}_{local}$ (Eq.~\ref{eq14}), after adding it to the baseline configuration, a $2.12\%$ performance gain is achieved in terms of average accuracy (\emph{3rd} row), which demonstrates the effectiveness of $\mathcal{L}_{local}$ on promoting the separability of feature representations. In addition, when $\mathcal{L}_{global}$ and $\mathcal{L}_{local}$ are simultaneously applied, the highest classification accuracy is obtained (\emph{4th} row), which shows the complementarity of global and local alignment constraints. 


\begin{table}[t]
        \begin{spacing}{1.38}
		\centering
		\caption{The performance of MRF-MSDA under three model configurations on \emph{Digits-five}.} \label{table_ablation_mrf}
	    \footnotesize
		\setlength{\tabcolsep}{1.6mm}
			\begin{tabular}{cc|ccccc|c}
			\toprule[1.0pt]

			 $\ \ \mathcal{L}_{cls}  \ $ & $\mathcal{L}_{MLE}$ & $\rightarrow$ \textbf{mm} & $\rightarrow$ \textbf{mt} &$\rightarrow$ \textbf{up} & $\rightarrow$ \textbf{sv} & $\rightarrow$ \textbf{syn}  & Avg \\ \specialrule{0em}{0pt}{0.8pt}
			
			\hline
			\ \ $ \checkmark$ &   & 86.54 & 99.07 & 98.36 & 82.90 & 92.80 & 91.93 \\
			& $\checkmark$ & 82.30 & 99.05 & 98.21 & 83.66 & 92.34 & 91.11 \\
			\ \ $\checkmark$  & $ \ \checkmark$ & \textbf{90.69} & \textbf{99.24} & \textbf{98.50} & \textbf{85.61} & \textbf{94.66} & \textbf{93.74} \\
			\bottomrule[1.0pt]		
			\end{tabular}
			\end{spacing}
\end{table}


\subsubsection{Ablation Study for MRF-MSDA} \label{sec6_1_2}

This set of experiments study the effect of classification and MLE-based objective functions on the MRF-MSDA model. Tab.~\ref{table_ablation_mrf} reports the performance of MRF-MSDA under three configurations on the Digits-five data set. When the classification constraint $\mathcal{L}_{cls}$ (Eq.~\ref{eq21}) or MLE-based constraint $\mathcal{L}_{MLE}$ (Eq.~\ref{eq22}) is individually applied (\emph{1st}/\emph{2nd} row), the classification accuracy is obviously lower than the full model configuration (\emph{3rd} row), \emph{i.e.} using both objective functions. These results illustrate the benefits of both joint distribution modeling and discriminative modeling on the observations. Through combining these two objectives, MRF-MSDA can derive more precise label prediction for the query sample.


\begin{figure*}[t]
	\centering
	\includegraphics[width=0.95\textwidth]{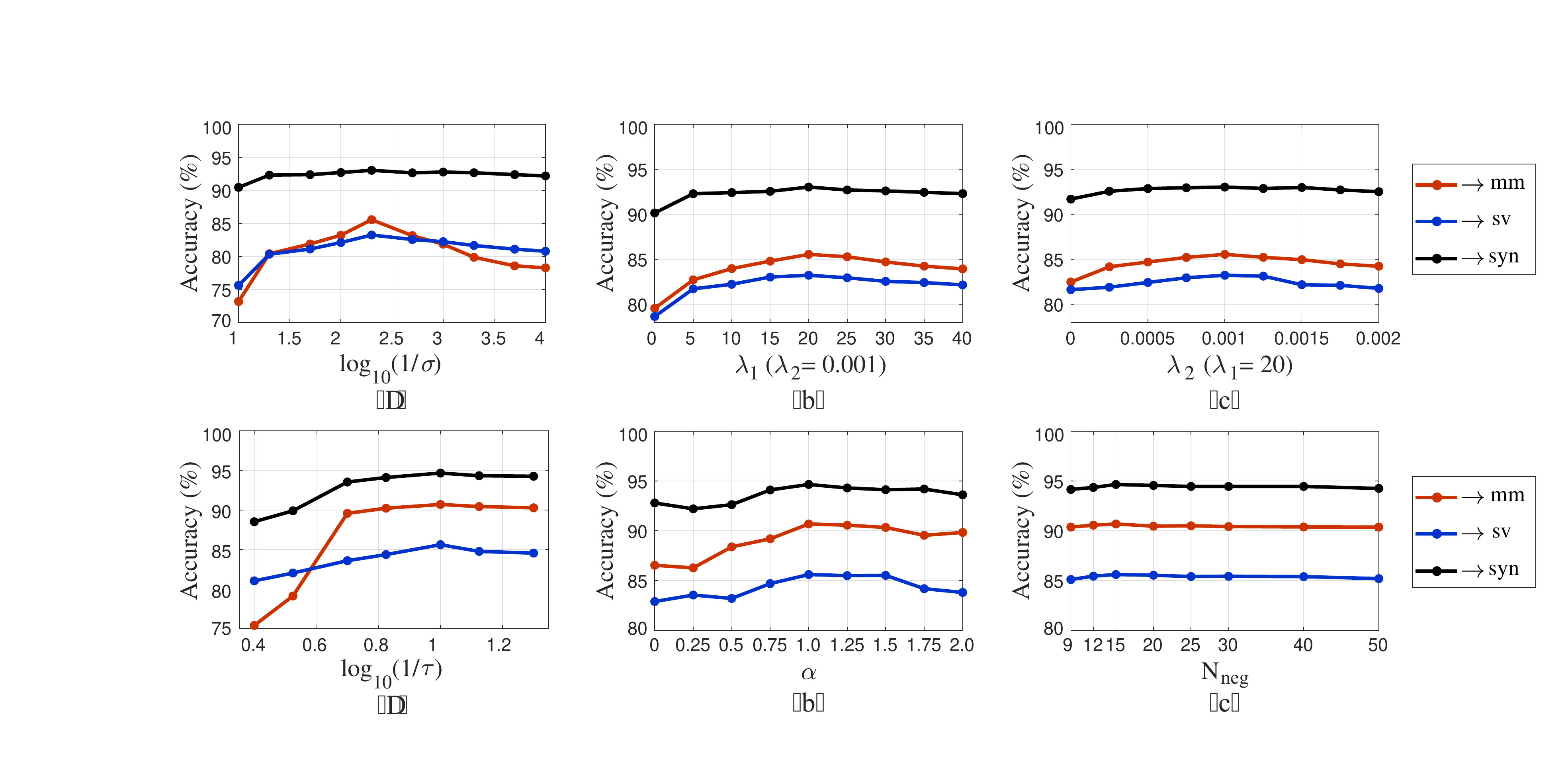}
	\caption{Sensitivity analysis for three parameters of CRF-MSDA on the \emph{Digits-five} data set.} 
	\vspace{-3mm}
	\label{fig_sensitivity_crf}
\end{figure*}


\begin{figure*}[t]
	\centering
	\includegraphics[width=0.95\textwidth]{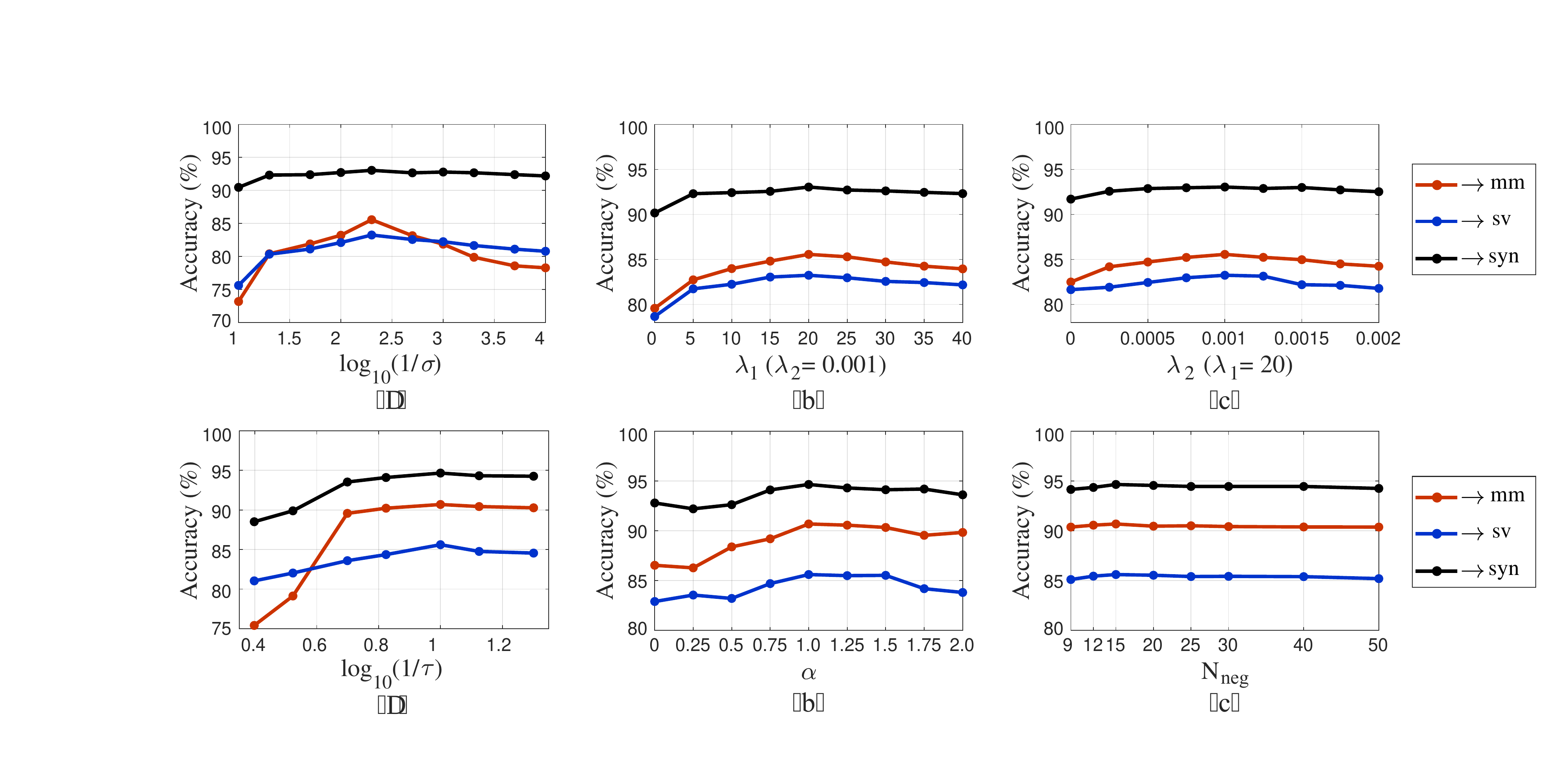}
	\caption{Sensitivity analysis for three parameters of MRF-MSDA on the \emph{Digits-five} data set.} 
	\vspace{-5mm}
	\label{fig_sensitivity_mrf}
\end{figure*}


\subsection{Sensitivity Analysis} \label{sec6_2}


\subsubsection{Sensitivity Analysis for CRF-MSDA} \label{sec6_2_1}

\textbf{Sensitivity of bandwidth parameter $\sigma$.} In this experiment, we discuss the selection of bandwidth parameter $\sigma$ which controls the sparsity of the adjacency matrix $\mathbf{A}$ defined in Eq.~\ref{eq4}. In Fig.~\ref{fig_sensitivity_crf}(a), we plot the performance of models trained with different $\sigma$ values. We can observe that, on all three tasks, the highest accuracy is achieved when the value of $\sigma$ is around 0.005. Under such condition, the adjacency matrix can capture the relations among observations most appropriately. Also, it is worth noticing that performance decay occurs when the adjacency matrix is too dense or sparse, \emph{i.e.} $\sigma > 0.05$ or $\sigma < 0.0005$. 

\textbf{Sensitivity of trade-off parameters $\lambda_1$, $\lambda_2$.} In this part, we evaluate the CRF-MSDA model's sensitivity to $\lambda_1$ and $\lambda_2$ which balance between different learning objectives. Fig.~\ref{fig_sensitivity_crf}(b) and Fig.~\ref{fig_sensitivity_crf}(c) show model's performance under various $\lambda_1$ ($\lambda_2$) values when the other trade-off parameter $\lambda_2$ ($\lambda_1$) is fixed. It can be observed that CRF-MSDA model's performance is not sensitive to $\lambda_1$ and $\lambda_2$ when they are around 20 and 0.001, respectively. When these two parameters approach 0, obvious performance decrease occurs, which again verifies that both global and local alignment constraints are indispensable. 




\begin{figure}[t]
	\centering
	\includegraphics[width=.98\linewidth]{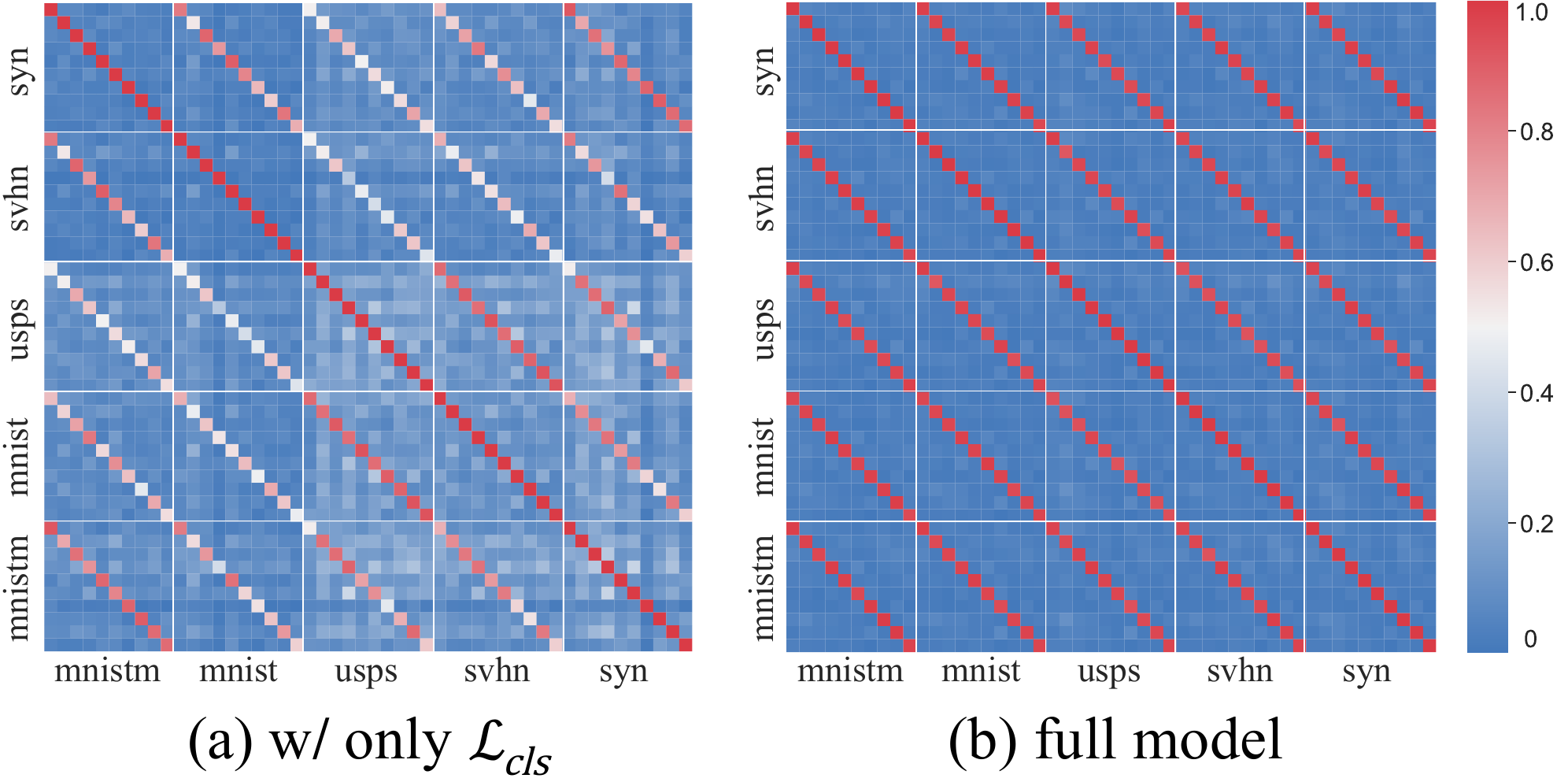}
	\caption{The adjacency matrix in CRF-MSDA. (Results are evaluated on the ``$\rightarrow$ \textbf{mm}'' task of \emph{Digits-five}.)} 
	\label{fig_visualization_crf}
	\vspace{-2mm}
\end{figure}


\begin{table}[t]
        \begin{spacing}{1.1}
		\centering
		\caption{Running time over 100 iterations of different methods on \emph{Digits-five} in both training and inference phase.} \label{table_ablation_time}
		\setlength{\tabcolsep}{3.2mm}
			\begin{tabular}{c|c|c}
			\toprule[1.0pt]

			 Methods & training (s)  & inference (s) \\ \specialrule{0em}{0pt}{0.8pt}
			\hline
			$\rm M^{3}SDA$~\cite{M3SDA} & 19.92 $\pm$ 0.07 & \textbf{4.96} $\pm$ 0.08 \\
			CRF-MSDA  & 17.41 $\pm$ 0.09 & 5.62 $\pm$ 0.12  \\
			MRF-MSDA & \textbf{16.54} $\pm$ 0.10 & 5.40 $\pm$ 0.08  \\
			\bottomrule[1.0pt]		
			\end{tabular}
			\end{spacing}
\end{table}


\subsubsection{Sensitivity Analysis for MRF-MSDA} \label{sec6_2_2}

\textbf{Sensitivity of temperature parameter $\tau$.} This experiments studies the selection of temperature parameter $\tau$ which scales the energy function defined in Eq.~\ref{eq17}. According to Fig.~\ref{fig_sensitivity_mrf}(a), when $\tau$ is around 0.1, the corresponding scaling can benefit the MRF-MSDA model to the greatest extent. With the increase of the temperature parameter, the model's performance drops apparently, \emph{e.g.} a nearly $15\%$ decrease on the ``$\rightarrow$ \textbf{mm}'' task when $\tau = 0.4$. This phenomenon illustrates that the joint distribution modeling of MRF-MSDA relies on a proper scale of energies to define the joint likelihoods. 


\textbf{Sensitivity of trade-off parameter $\alpha$.} In this part, we analyze the sensitivity of the trade-off parameter $\alpha$ which balances between the objectives for classification and maximum likelihood estimation. From the line chart in Fig.~\ref{fig_sensitivity_mrf}(b), we can observe that the MRF-MSDA model performs stably better when the value of $\alpha$ is around $1.0$ compared to other settings. Such value of $\alpha$ is able to attain an appropriate balance between two learning objectives.


\textbf{Sensitivity of negative sampling size $N_{neg}$.} The optimization of MRF-MSDA depends on sampling negative Markov networks to contrast with, which derives a parameter of negative sampling size $N_{neg}$. Based on the results shown in Fig.~\ref{fig_sensitivity_mrf}(c), we can conclude that the performance of MRF-MSDA is not sensitive to the value of $N_{neg}$, which, we think, owes to the strong negative samples (\emph{i.e.} negative networks with only minor difference relative to the positive one) used in our method.


\begin{figure*}[t]
	\centering
	\includegraphics[width=0.9\textwidth]{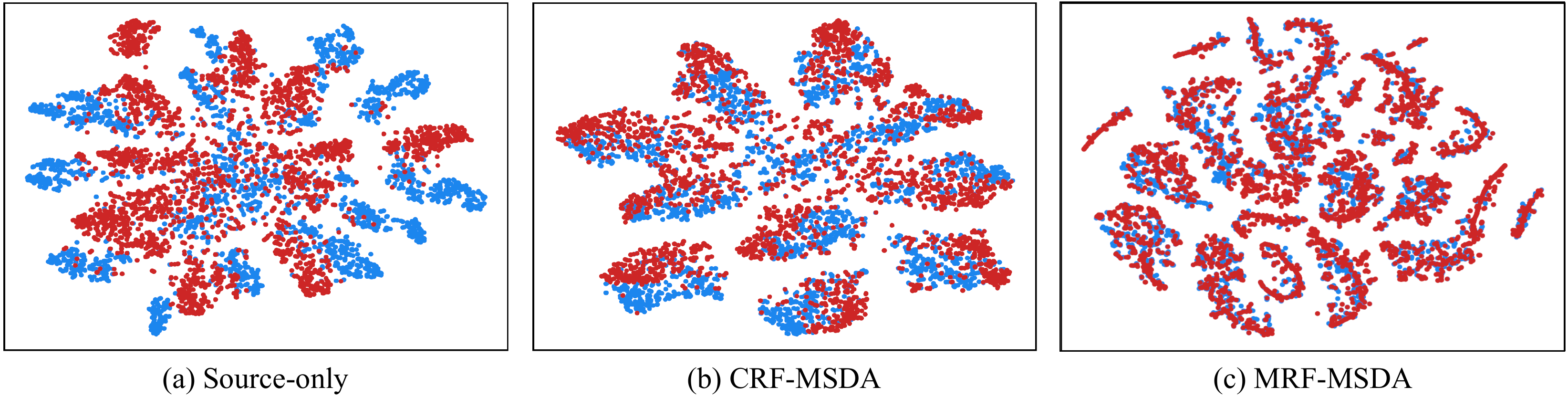}
	\caption{Visualization of feature embeddings. (All results are evaluated on the ``$\rightarrow$ \textbf{mm}'' task.)} 
	\vspace{-5mm}
	\label{fig_tsne}
\end{figure*}


\begin{figure}[t]
	\centering
	\includegraphics[width=.94\linewidth]{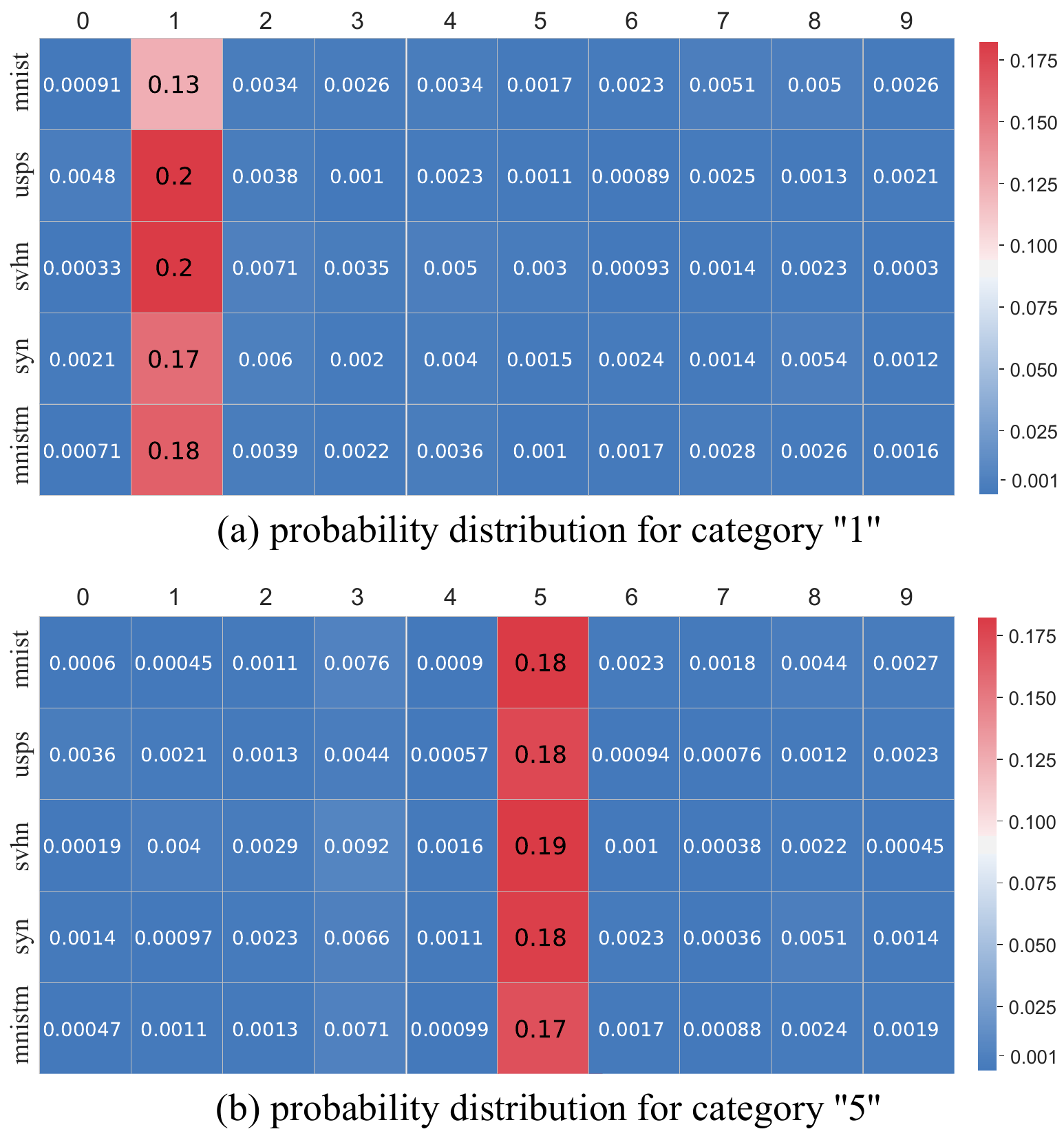}
	\caption{The probability distribution in MRF-MSDA. (Results are evaluated on the ``$\rightarrow$ \textbf{mm}'' task of \emph{Digits-five}.)} 
	\label{fig_visualization_mrf}
	\vspace{-2mm}
\end{figure}


\subsection{Times Complexity Analysis} \label{sec6_3}
Table~\ref{table_ablation_time} reports the running time of 100 iterations in both training and inference phase of different methods on Digits-five data set. The hardware conditions for the experiments are Intel(R) Xeon(R) CPU E5-2620 v4@2.40 GHz with 8 processors and one NVIDIA TITAN Xp GPU. All the reported results are averaged over 10 independent runs under the same configuration. From the table we can observe that in the training phase, under 100 iterations, the MRF model is about 0.9 seconds faster than the CRF model. And in the inference phase, MRF-MSDA model also runs slightly faster than CRF-MSDA model. These experimental results verify that MRF-MSDA is indeed more computationally efficient than CRF-MSDA for both learning and inference. 


\subsection{Visualization} \label{sec6_4}


\subsubsection{Visualization for CRF-MSDA} \label{sec6_4_1}

In the CRF-MSDA model, the adjacency matrix $\mathbf{A}$ (Eq.~\ref{eq4}) quantifies the category-level relevance between various domains. In Fig.~\ref{fig_visualization_crf}, we visualize $\mathbf{A}$ under two model configurations, in which each pixel denotes the adjacency between two categories from arbitrary domains. Compared to the configuration with only classification constraint, the full model applying both classification and alignment objective functions achieves better cross-domain consistency on the relevance among various categories, which illustrates the effectiveness of global-level alignment. 



\subsubsection{Visualization for MRF-MSDA} \label{sec6_4_2}

In this part, we visualize the category-specific probability distribution derived by MRF-MSDA. For the query sample $q$, we combine the probabilities $p(y_d = m, y = k | q)$ ($1 \leqslant m \leqslant M+1$, $1 \leqslant k \leqslant K$) defined in Eq.~\ref{eq18} as a probability matrix $\mathbf{P}_q \in \mathbb{R}^{(M+1) \times K}$. Through averaging $\mathbf{P}_q$ over all the query samples within a specific category, we can obtain the probability matrix for that category. In Fig.~\ref{fig_visualization_mrf}, we visualize the probability matrix for category ``1'' and ``5'' on the target domain's test set of ``$\rightarrow$ \textbf{mm}'' task. We can observe that high probability values evenly distribute on the corresponding categories (``1'' or ``5'') of various domains, which demonstrates that MRF-MSDA effectively aligns the samples within the same category and separates the ones from distinct categories in the latent space.  

\subsubsection{Visualization of feature embeddings} \label{sec6_4_3}
In Figure \ref{fig_tsne}, we utilize t-SNE \cite{tsne} to visualize the feature distributions of one of source domains (SVHN) and target domain (MNIST-M). Compared with the Source-only baseline, the proposed CRF-MSDA and MRF-MSDA model make the features of target domain more discriminative and better aligned with those of source domain. Compared to CRF-MSDA, the feature representations derived by MRF-MSDA are better aligned across two domains, which is in line with the better empirical performance of MRF-MSDA on the ``$\rightarrow$ \textbf{mm}'' task.


\section{Conclusions and Future Work} \label{sec7}

In this work, we aim to address the Multi-Source Domain Adaptation (MSDA) problem. Specifically, we propose two graphical models, \emph{i.e.} Conditional Random Field for MSDA (CRF-MSDA) and Markov Random Field for MSDA (MRF-MSDA), to realize cross-domain joint modeling and learnable domain combination. Extensive experiments on various benchmark data sets of MSDA illustrate the superior performance of our methods over existing works. In the future work, we will explore other graphical models for MSDA, \emph{e.g.} Bayesian networks and chain graphs.








\ifCLASSOPTIONcompsoc
  \section*{Acknowledgments}
  This work was supported by National Science Foundation of China (U20B2072, 61976137). This work was also partially supported by Grant YG2021ZD18 from Shanghai Jiao Tong University Medical Engineering Cross Research.
\else
  \section*{Acknowledgment}
\fi

\ifCLASSOPTIONcaptionsoff
  \newpage
\fi


\bibliographystyle{IEEEtran}
\bibliography{reference}

\end{document}